\documentclass[journal]{IEEEtran}
\ifCLASSINFOpdf
  \usepackage{graphicx}
\else
\fi
\usepackage{amsmath}
\usepackage{array}
\ifCLASSOPTIONcompsoc
  \usepackage[caption=false,font=normalsize,labelfont=sf,textfont=sf]{subfig}
\else
  \usepackage[caption=false,font=footnotesize]{subfig}
\fi

\ifCLASSOPTIONcaptionsoff
  \usepackage[nomarkers]{endfloat}
 \let\MYoriglatexcaption\caption
 \renewcommand{\caption}[2][\relax]{\MYoriglatexcaption[#2]{#2}}
\fi
\usepackage{url}
\usepackage{amssymb}
\usepackage{booktabs}
\usepackage{xcolor}         % colors

\usepackage{algorithm}
\usepackage{algorithmic}
\usepackage{multirow}

\begin{document}

\title{Enabling Population-Based Architectures for\\ Neural Combinatorial Optimization}


\author{Andoni~I.~Garmendia, Josu~Ceberio, and Alexander~Mendiburu,%
\thanks{The authors are with the Intelligent Systems Group, University of the Basque Country (UPV/EHU), San Sebastian, 20018 Gipuzkoa, Spain (e-mail: andoni.irazusta@ehu.eus).}%
\thanks{A. I. Garmendia, J. Ceberio, and A. Mendiburu, “Enabling Population-Based Architectures for Neural Combinatorial Optimization,” IEEE Transactions on Evolutionary Computation, 2026, doi: 10.1109/TEVC.2026.3730516.}%
\thanks{© 2026 IEEE.  Personal use of this material is permitted.  Permission from IEEE must be obtained for all other uses, in any current or future media, including reprinting/republishing this material for advertising or promotional purposes, creating new collective works, for resale or redistribution to servers or lists, or reuse of any copyrighted component of this work in other works.}
}

\maketitle

\begin{abstract}

Neural Combinatorial Optimization (NCO) has mostly focused on learning policies, typically neural networks, that operate on a single candidate solution at a time, either by constructing one from scratch or iteratively improving it.
In contrast, decades of work in metaheuristics have shown that maintaining and evolving populations of solutions improves robustness and exploration, and often leads to stronger performance. 
To close this gap, we study how to make NCO explicitly population-based by learning policies that act on sets of candidate solutions. We first propose a simple taxonomy of “population awareness” levels and use it to highlight two key design challenges: (i) how to represent a whole population inside a neural network, and (ii) how to learn population dynamics that balance intensification (generating good solutions) and diversification (maintaining variety).
We make these ideas concrete with two complementary tools: one that improves existing solutions using information shared across the whole population, and the other generates new candidate solutions that explicitly balance being high-quality with diversity.
Experimental results on Maximum Cut and Maximum Independent Set indicate that incorporating population structure is advantageous for learned optimization methods and opens new connections between NCO and classical population-based search.

\end{abstract}


\ifCLASSOPTIONpeerreview
\begin{center} \bfseries EDICS Category: 3-BBND \end{center}
\fi
\IEEEpeerreviewmaketitle

\section{Introduction}
\label{sec:intro}

\IEEEPARstart{N}{eural} Combinatorial Optimization (NCO)~\cite{bengio2021machine,mazyavkina2021reinforcement,bello2016neural} studies \emph{learning-based} approaches for solving combinatorial optimization problems of different nature, such as routing~\cite{kool2018attention}, assignment~\cite{wang2021neural}, scheduling~\cite{zhang2020learning}, or graph problems~\cite{barrett2020exploratory}. 
Instead of manually designing problem-specific heuristics, NCO trains parametric policies, typically neural networks, that either construct solutions step by step or improve existing solutions via learned local moves. 

NCO is still far less mature than the rich body of classical heuristics and metaheuristics~\cite{blum2003metaheuristics,liu2023good}, and has not yet consistently surpassed state-of-the-art methods in terms of solution quality and robustness across diverse benchmarks.
Nonetheless, compared with classical methods (e.g., greedy construction, local search, simulated annealing, tabu search, or evolutionary algorithms)~\cite{blum2003metaheuristics}, NCO offers three benefits that are increasingly relevant at scale.
First, it can automatically learn problem-specific guidance rules from data, for example, how to rank candidate moves, which neighborhoods to try first, or how to extend partial solutions~\cite{bengio2021machine,mazyavkina2021reinforcement,khalil2017learning}.
Second, once trained, the resulting policy can be reused across many new instances with minimal additional computation or per-instance tuning~\cite{kool2018attention,kwon2020pomo}.
Third, NCO maps naturally to modern accelerators (GPUs), enabling batched inference and high-throughput evaluation of candidate moves or solutions~\cite{kool2018attention,kwon2020pomo}.

In practice, NCO and metaheuristics can coexist with different deployment profiles: metaheuristics are typically easier to prototype and broadly robust, whereas learned policies require more engineering and data but excel when rapid, repeated inference over many instances is needed~\cite{liu2023good}, such as in online decision-making systems~\cite{luo2024self}, real-time applications~\cite{cappart2023combinatorial}, or large-scale optimization pipelines~\cite{zhou2024instance}.

Beyond the advantages outlined above, the development of NCO methods has largely mirrored the historical trajectory of heuristic and metaheuristic methods~\cite{blum2003metaheuristics}.
Early work adopted \emph{Neural Constructive} (NC) policies~\cite{vinyals2015pointer,bello2016neural,kool2018attention,kwon2020pomo} that build solutions iteratively, adding one item at a time using a learned heuristic. This was later followed by \emph{Neural Improvement} (NI)~\cite{chen2019learning,lu2019learning,barrett2020exploratory,wu2021learning}, echoing classical local search, where a policy proposes local moves to improve a given solution. More recent efforts introduce tabu-like memory mechanisms~\cite{barrett2022learning,garmendia2024marco,chalumeau2024memory}, which track recent action histories to prevent getting stuck in local optima.

However, to the best of our knowledge, there is no prior work that learns a population-aware policy in an NCO framework.
Instead, the most closely related lines of work fall into two categories:
(i) \textit{Deep Learning-assisted metaheuristics}, where learning parametrizes specific components of the metaheuristic while the population mechanics remain classical (e.g., learned crossover and selection~\cite{liu2023neurocrossover}, Genetic Algorithm initialization~\cite{greenberg2025accelerating}, or learning heuristics for Ant Colony Optimization ~\cite{ye2024deepaco,dorigo2019ant}),
(ii) \emph{Neural ensembles}, that train several different policies (or decoders) so that each one generates solutions in its own specialized way~\cite{grinsztajn2023winner,hottung2024polynet}. These approaches can produce diverse solutions, but the policies do not interact with one another: there is no population-level coordination, and each policy works as an isolated individual.

Given the long-standing success of population-based metaheuristics in combinatorial optimization~\cite{blum2003metaheuristics,vidal2014unified,dorigo2019ant}, and the strong improvements already demonstrated by neural constructive~\cite{kwon2020pomo,luo2023neural} and neural improvement approaches~\cite{barrett2020exploratory,wu2021learning}, extending NCO methods toward a neural population-based search paradigm represents a natural, highly promising and still underexplored direction.
However, this extension raises several nontrivial challenges: how to process and evolve a population within a neural architecture; how to balance intensification and diversification when the \emph{operators} are neural networks; and how to train these systems efficiently, as the shift from single-solution optimization to population evolution introduces complex interactions that complicate the learning signal.

In this work, we take a first step in that direction by framing neural population-based search as the problem of learning \emph{population operators} that act on sets of solutions rather than isolated individuals.
We identify representation, coordination, and training as the core design dimensions, and instantiate them through two complementary components: 
i) a contextual improvement policy, a learned local search operator that uses a central memory to share information throughout the population, and 
(ii) a conditioned constructive policy, an operator that generates new solutions, using a steerable exploration weight to explicitly regulate the trade-off between solution quality and population diversity.
Combining these components yields a population-based NCO framework (PB-NCO) that alternates memory-guided local search with diversity-aware \emph{restarts}, where non-improving individuals are discarded and replaced by new candidates sampled from the conditioned constructive policy.

We assess the framework on Maximum Cut (MC)~\cite{dunning2018works} and Maximum Independent Set (MIS)~\cite{lawler1980generating} across different graph families and problem sizes. The study includes performance and diversity analyses, as well as ablations that quantify the sensitivity of key components. We compare against classical population-based methods and leading neural baselines under matched time budgets. Overall, the proposed population-based methods achieve promising results that are competitive with these strong baselines.

The remainder of this paper is organized as follows.
Section~\ref{background} reviews relevant background on NCO, including search settings, policy architectures, and learning paradigms.
Section~\ref{sec:design_challenges} introduces a taxonomy of different population-awareness levels and formalizes the main design challenges for neural population-based methods.
Section~\ref{sec:method} presents our proposals, detailing the contextual neural improvement (cNI) and conditioned neural constructive (cNC) components and their integration into the complete framework PB-NCO.
Section~\ref{sec:benchmark_problems} defines the benchmark problems considered in this study.
Section~\ref{sec:exp_protocol} describes the experimental protocol, and Section~\ref{sec:results} reports the empirical results.
Section~\ref{sec:discussion} discusses implications, limitations, and natural extensions of our approach, and Section~\ref{sec:conclusion} concludes the paper.

\section{Background}
\label{background}

\subsection{Combinatorial Optimization}
\label{subsec:CO_setup}

Let $\mathcal{I}$ denote an instance of a combinatorial optimization problem with a feasible set 
$\mathcal{S}(\mathcal{I})$ and an objective function 
$f_{\mathcal{I}} : \mathcal{S}(\mathcal{I}) \to \mathbb{R}$. 
Without loss of generality, we adopt the maximization convention 
(minimization problems can be handled by sign inversion). 
Each feasible configuration $\mathbf{s} \in \mathcal{S}(\mathcal{I})$ represents a candidate solution, 
and the goal is to find one that maximizes the objective value:
\begin{equation}
    \mathbf{s}^* \in \arg\max_{\mathbf{s} \in \mathcal{S}(\mathcal{I})} f_{\mathcal{I}}(\mathbf{s}).
\end{equation}

\subsection{Neural Combinatorial Optimization}
\label{subsec:setup}

Neural Combinatorial Optimization (NCO)~\cite{bengio2021machine,mazyavkina2021reinforcement,bello2016neural}
refers to methods that solve combinatorial problems by learning a policy directly from data.
Formally, we consider a state space $\mathcal{X}$ and, for each state $x \in \mathcal{X}$, a set of feasible actions $\mathcal{A}(x)$.
A policy $\pi_\theta$ with parameters $\theta$ maps each state to a probability distribution over actions $\pi_\theta : \mathcal{X} \to \mathcal{A}$.

NCO typically follows a \emph{train–inference} scheme: in the training phase, $\pi_\theta$ is optimized offline on a distribution of problem instances to discover decision patterns that lead to high-quality solutions; at inference time, the same policy is applied to new unseen instances, requiring only forward passes of the network and thus incurring a relatively low computational cost compared to training.

In NCO, the search process is modeled as a sequence of decisions.
At each step $t$, the policy $\pi_\theta$ observes the current state $x_t \in \mathcal{X}$, which encodes both the problem instance $\mathcal{I}$ and its current search context, and selects an action $a_t \in \mathcal{A}(x_t)$.
This interaction produces a trajectory
\[
    \tau = (x_0, a_0, x_1, a_1, \ldots, a_T, x_T).
\]

Two settings are typically distinguished. In the \textbf{Constructive setting}, the state $x_t$ encodes a partial solution,
$x_t = (\mathcal{I}, \text{partial } s_t)$, typically starting from an empty solution.
At each step, the policy selects an action that adds one component, and the process continues until a complete feasible solution $s_T \in \mathcal{S}(\mathcal{I})$ is obtained.
In the \textbf{Improvement setting,} the state contains a complete feasible solution,
$x_t = (\mathcal{I}, s_t)$, and actions correspond to local modifications within a neighborhood
$\mathcal{N}(s_t) \subseteq \mathcal{S}(\mathcal{I})$ that aim to improve it.
These moves are applied iteratively until a stopping criterion is met (e.g., no further improvement, iteration limit, or time budget).

In both settings, the policy $\pi_\theta$ must encode the combinatorial structure of the instance together
with the current search state and map this representation to an action.
The concrete way this is implemented is detailed in the following subsection.

\subsection{Model Architecture}
\label{subsec:nn_arch}

\begin{figure}[!h]
\centering
\includegraphics[width=0.495\textwidth]{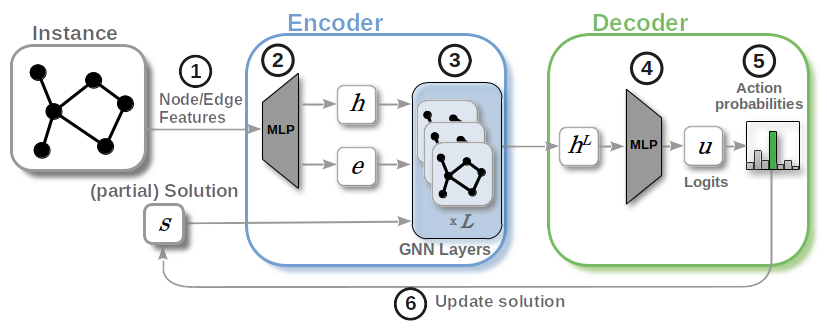}
\caption{
High-level encoder--decoder architecture used in Neural Combinatorial Optimization. First, graph features associated with nodes and edges are extracted from the problem instance and the current search state (1). These raw features are then projected into a common $d$-dimensional latent space, producing initial node embeddings $h^{(0)}$ and edge embeddings $e^{(0)}$ (2). The embeddings are processed by a GNN through $L$ layers (3). The final embeddings are used by a decoder (4) to produce action logits, which define action probabilities (5) used to modify or construct the solution (6).}
\label{fig:nco_architecture}
\end{figure}

Most NCO policy architectures follow a two-stage design (see Fig.~\ref{fig:nco_architecture}): 
(i) an \emph{encoder} that receives the instance features together with the current state of the search and produces a representation in the latent space, and 
(ii) a \emph{decoder} that uses these representations to select an action specifying how to construct or modify the solution.

The encoder typically operates on graph-structured inputs, as many combinatorial optimization problems can be naturally represented as a graph $\mathcal{G}=(\mathcal{V},\mathcal{E})$~\cite{cappart2023combinatorial}, where nodes denote items or decisions and edges encode pairwise relations (e.g., distances, conflicts, compatibilities). 
Each node and each edge is described by two types of features: \emph{static} features, which encode instance-specific information such as weights or costs and do not change throughout the search process, and \emph{dynamic} features, which encode the current search state, and are updated as the search progresses, for example, to indicate the current (partial or complete) solution, the role of a node in the neighborhood of the current candidate, or other problem-specific indicators.

All static and dynamic node and edge features (step 1 in Fig.~\ref{fig:nco_architecture}) are first projected into a $d$-dimensional embedding space, yielding initial node embeddings $h_i^{(0)} \in \mathbb{R}^d$ for each node $i \in \mathcal{V}$ and, similarly, initial edge embeddings $e_{ij}^{(0)} \in \mathbb{R}^d$ for each edge $(i,j) \in \mathcal{E}$ (step 2).
These embeddings are then processed by a Graph Neural Network (GNN)~\cite{cappart2023combinatorial} for $L$ message-passing layers (step~3)\footnote{GNNs are a standard choice for the encoder because they respect the symmetries of combinatorial problems (permutation equivariance with respect to node order) and allow the same architecture to operate on graphs of varying sizes.}, producing final node embeddings $h_i^{(L)} \in \mathbb{R}^d, \qquad i \in \mathcal{V}$, which are subsequently inputted to the decoder.

The decoder (typically a multi-layer perceptron) produces one scalar action logit for each feasible action.
Depending on the problem and the search setting (constructive or improvement), these actions are defined either on nodes or on edges.
For node-based actions, it maps each node embedding $h_i^{(L)} \in \mathbb{R}^d$ to a logit $u_i \in \mathbb{R}$ (step~4), representing the relative preference for selecting node $i$.
For edge-based actions (e.g., pairwise operators), we first form an edge embedding by concatenating the endpoint embeddings, $h_{ij} = [h_i^{(L)}; h_j^{(L)}] \in \mathbb{R}^{2d}$, and apply the same decoder to obtain an edge logit $u_{ij} \in \mathbb{R}$.
A softmax over the feasible action set $\mathcal{A}(x_t)$ then converts these logits into a probability distribution over actions (step~5):
\begin{equation}
\pi_\theta(a_t = a \mid x_t) =
\frac{\exp(u_a)}{\sum_{a' \in \mathcal{A}(x_t)} \exp(u_{a'})},
\label{eq:decoder-softmax}
\end{equation}
from which an action is sampled or greedily selected to update the current solution (step~6).

For a comprehensive overview of GNN-based architectures for combinatorial optimization, including message passing and action decoding mechanisms, we refer the reader to Cappart et al.~\cite{cappart2023combinatorial}.

\subsection{Policy Learning}
\label{sec:train-signal}

The original NCO approaches~\cite{bello2016neural} were developed within a Reinforcement Learning (RL)~\cite{sutton2018reinforcement} framework. 
Although earlier~\cite{vinyals2015pointer} and subsequent studies~\cite{bengio2021machine} have explored supervised and unsupervised training schemes, RL, and in particular policy-gradient methods~\cite{mazyavkina2021reinforcement}, remain the predominant learning paradigm in the NCO literature.

In this setting, the policy $\pi_\theta$ is executed on many instances, receives feedback in the form of rewards, and its parameters $\theta$ are updated so as to maximize the expected return, increasing the probability of actions that yielded higher returns.
Formally, let $r_t$ denote the reward given at decision step $t$, and define the discounted return from time $t$ as
\begin{equation}
\label{eq:g_t}
\hat G_t
=
\sum_{t' = t}^{T} \gamma^{\,t'-t} r_{t'},
\qquad \gamma \in [0,1],
\end{equation}
where the discount factor $\gamma$ can be used to emphasize earlier steps and reduce variance in long decision-trajectories.

In the policy-gradient (PG) scheme, the objective function is
\begin{equation}
\label{eq:pg}
\mathcal{J}_{\mathrm{PG}}(\theta)
=
\mathbb{E}_{\tau\sim\pi_\theta}\!\Bigg[\sum_{t=0}^{T}
\log \pi_\theta(a_t\mid x_t)\,A_t\Bigg],
\end{equation}
where $A_t = \hat G_t - B(x_t)$ is the \emph{advantage}, measuring how much better an action performs compared to an action-independent baseline $B(x_t)$.
The baseline acts as a control variate: it centers the returns and reduces gradient variance, without changing which policy is optimal.
Intuitively, Equation~\eqref{eq:pg} increases the probability of actions with positive $A_t$ and decreases it for negative $A_t$.
This formulation applies to both the constructive and improvement settings; only the definition of the reward $r_t$ in Equation~\eqref{eq:g_t} differs.

In the \emph{constructive setting}, rewards are only given once a complete solution ($s_T$) is constructed: all intermediate rewards are zero and the terminal reward equals the objective:
\begin{equation}
\label{eq:nc_reward}
r_t = 0 \quad (t<T), 
\qquad
r_T = f_{\mathcal{I}}(s_T).
\end{equation}
In this purely terminal case, $\gamma$ has no effect.

In the \emph{improvement setting}, the reward is defined per improvement step as the immediate gain in objective value:
\begin{equation}
\label{eq:ni_reward}
r_t \;=\; f_{\mathcal{I}}(s_{t+1}) - f_{\mathcal{I}}(s_t)
\end{equation}
Here, $\gamma=1$ yields undiscounted finite-horizon optimization; $0<\gamma<1$ progressively downweights later improvements (more myopic), and $\gamma=0$ reduces updates to purely one-step improvement.

Once the reward values are computed, policy parameters $\theta$ are updated using stochastic gradient ascent over the objective $\mathcal{J}_{\mathrm{PG}}$:
\begin{equation}
\label{eq_reinforce}
\theta \;\leftarrow\; \theta \;+\; \eta\,\widehat{\nabla_\theta \mathcal{J}_{\mathrm{PG}}}
\end{equation}
where $\eta>0$ is the learning rate and $\widehat{\nabla_\theta \mathcal{J}_{\mathrm{PG}}}$ denotes the gradient estimate computed from the sampled trajectories.

\section{Taxonomy and Design Challenges for Neural Population-based Methods}
\label{sec:design_challenges}

\begin{figure*}[t]
    \centering
    \subfloat[]{%
        \includegraphics[width=0.32\textwidth]{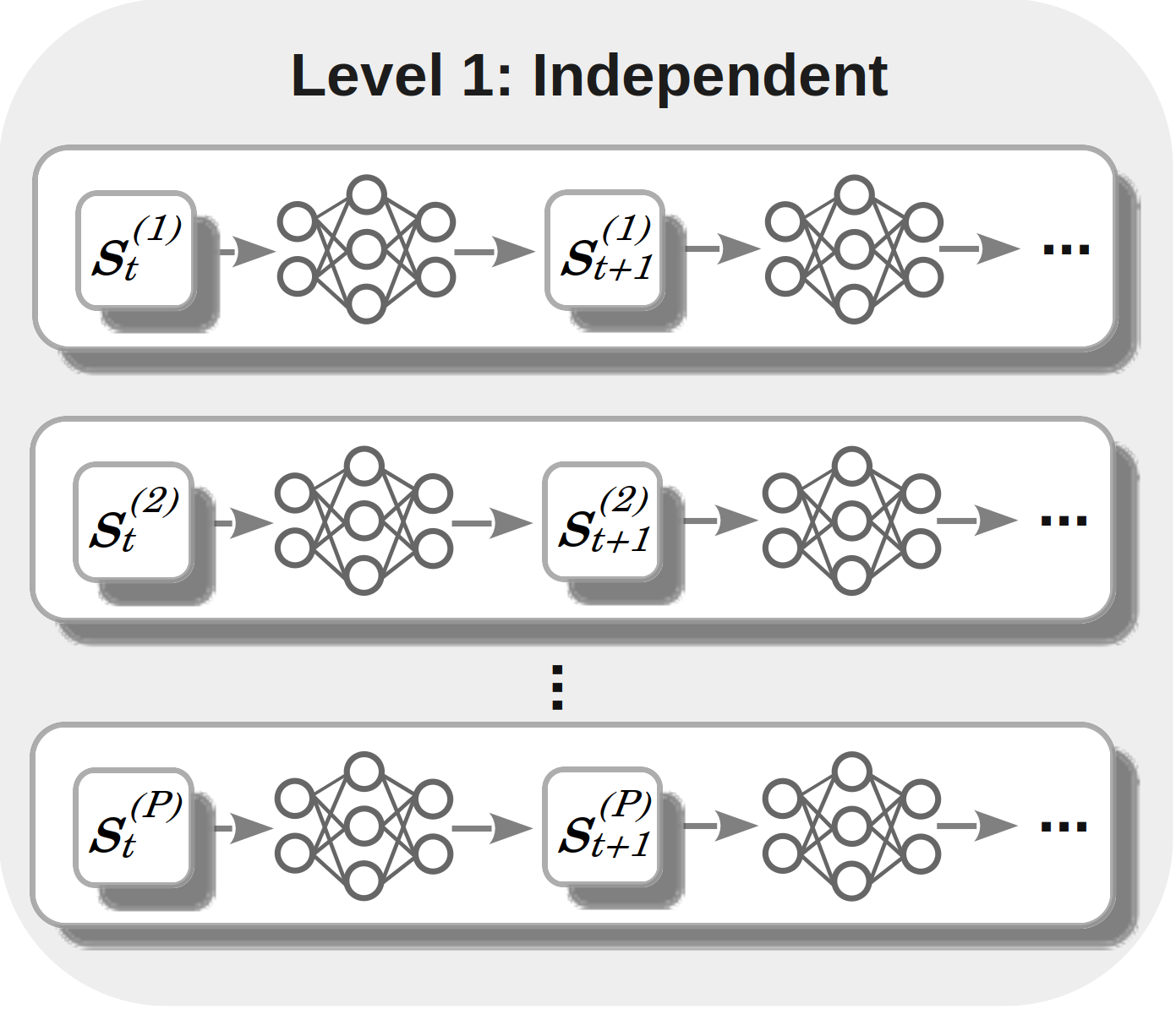}%
    }\hfill
    \subfloat[]{%
        \includegraphics[width=0.32\textwidth]{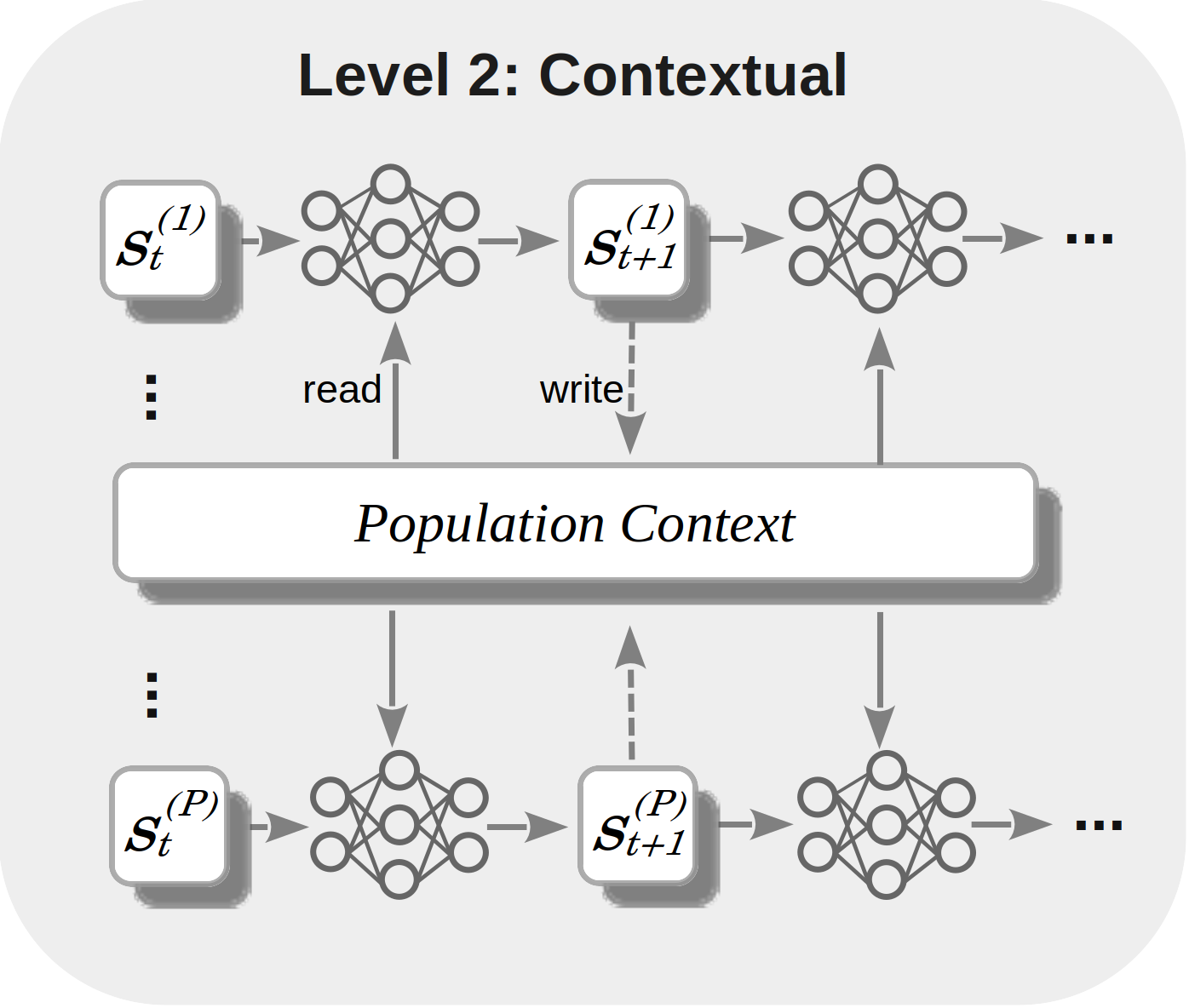}%
    }\hfill
    \subfloat[]{%
        \includegraphics[width=0.32\textwidth]{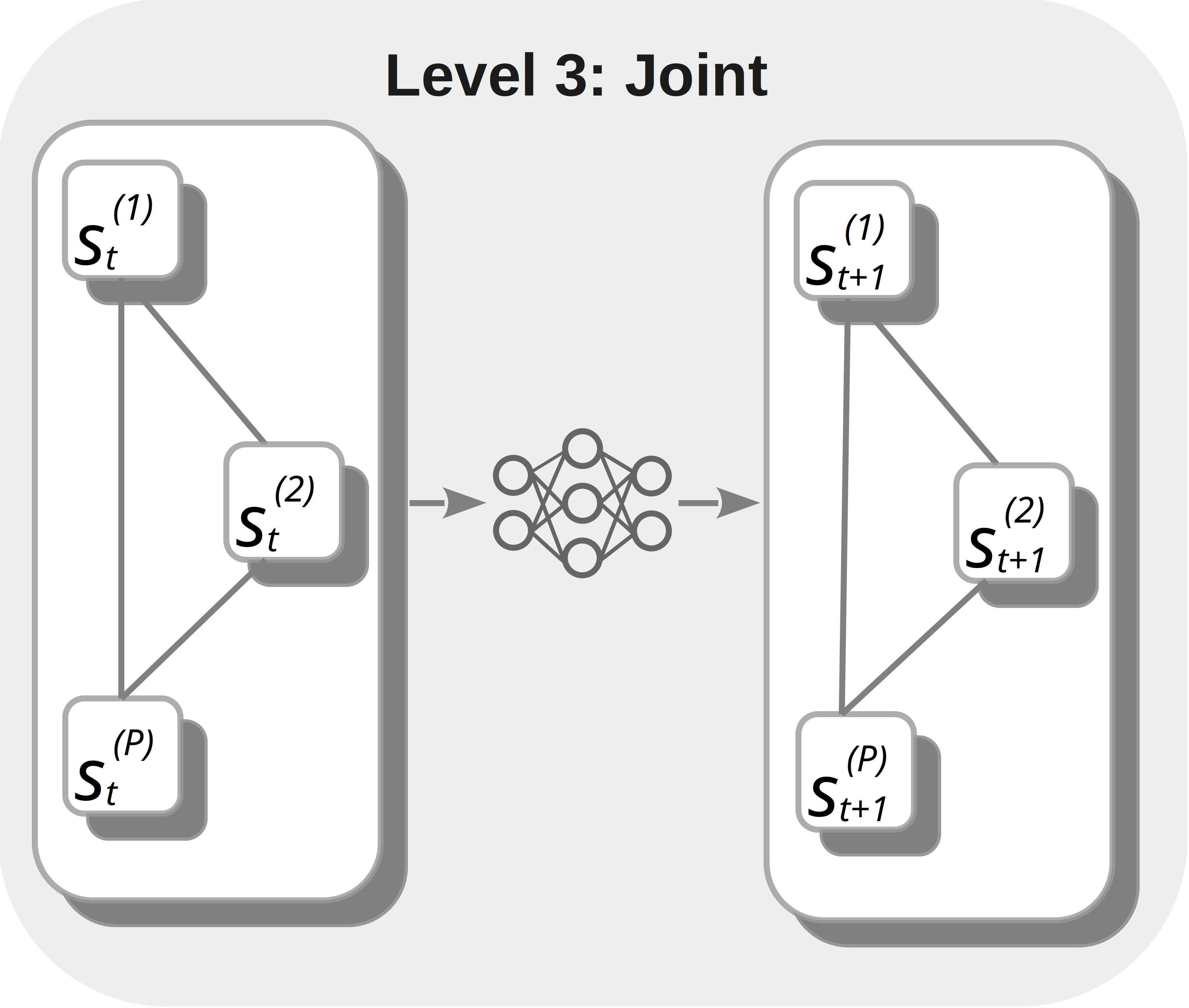}%
    }
    \caption{
        \textbf{Three levels of population awareness in neural population-based methods.}
        \textbf{(a) Independent:} Each solution $s_t^{(i)}$ is improved in isolation. The policy is applied in parallel to multiple starting points, but there is no exchange of information between runs (equivalent to multi-start NCO).
        \textbf{(b) Contextual:} Each solution is updated based on its own state and an external context signal. This context can be global (e.g., a shared memory) or directed (e.g., peer-to-peer communication from specific neighbors).
        \textbf{(c) Joint:} The model reasons jointly about multiple solutions at once, allowing explicit pairwise or group interactions across the population.}
    \label{fig:pop_methods}
\end{figure*}

As discussed in Section~\ref{sec:intro}, most existing NCO methods operate on a single solution at a time. 
Extending them to a population-based framework raises new questions about how neural architectures could effectively represent and evolve a \emph{set} of solutions.

We denote by $\mathcal{P}_t = \{ s_t^{(i)} \}_{i=1}^{P}$ a population of $P$ candidate solutions at iteration $t$ for an instance $\mathcal{I}$.
A neural population-based method can then be viewed as a learned operator
\begin{equation}
    F_\theta : (\mathcal{I}, \mathcal{P}_t) \mapsto \mathcal{P}_{t+1},
\end{equation}
whose goal is to transform $\mathcal{P}_t$ into a new population $\mathcal{P}_{t+1}$ such that solution quality improves over time.

To reason about potential population-based extensions of NCO, we introduce a simple taxonomy that classifies how much population information is exposed to each decision (see Fig.~\ref{fig:pop_methods}), ranging from no population awareness to full joint reasoning over multiple individuals.
This framework allows us to interpret existing single-solution NCO methods as a degenerate, population-unaware case, and to use the richer levels to motivate the specific design choices behind our cNI and cNC components.

\begin{itemize} 
\item \textbf{Level 1: Independent.} 
    The policy evolves each solution $s_t$ based solely on its own trajectory, with no information exchanged across runs.
    In practice, the $P$ solutions are treated as a batch: the same improvement policy is applied in parallel from multiple random starts, and the model only sees $P$ isolated trajectories, as in standard single-trajectory NCO methods~\cite{bengio2021machine,bello2016neural,kool2018attention}.
    Formally, the policy factorizes over solutions as
    \begin{equation}
    \label{eq:level1}
        x_t^{(i)} \;=\; (\mathcal{I}, s_t^{(i)}), 
        \qquad
        a_t^{(i)} \sim \pi_\theta\big(a \mid x_t^{(i)}\big),
    \end{equation}
    and the next population $\mathcal{P}_{t+1}$ is obtained by applying the underlying NCO update rule to each solution independently. %CITATION
    
\item \textbf{Level 2: Contextual.} 
    The policy acts locally on each individual, but it is conditioned on an external context signal that depends on the rest of the population.
    The search state is
    \begin{equation}
    \label{eq:level2-state}
        x_t^{(i)} = \big(\mathcal{I}, s_t^{(i)}, c_t^{(i)}\big),
    \end{equation}
    where
    \begin{equation}
    \label{eq:level2-context}
        c_t^{(i)} \;=\; C_\theta\big(s_t^{(i)}, \mathcal{P}_t\big)
    \end{equation}
    is a population-derived context vector for individual $i$.
    The mapping $C_\theta$ can be any function of the population and the focal individual, such as a shared summary over all solutions, a $k$-nearest-neighbor descriptor, or an aggregation over neighbors in a similarity graph.

\item \textbf{Level 3: Joint.}
    In the richest form of population awareness, the policy processes several solutions in a single joint state
    \begin{equation}
        X_t \;=\; \big(x_t^{(1)}, \dots, x_t^{(P')}\big),
        \qquad
        x_t^{(i)} = (\mathcal{I}, s_t^{(i)}),
    \end{equation}
    for some subset $\mathcal{P}_t' = \{s_t^{(i)}\}_{i=1}^{P'}$ of the population (e.g., the whole population or a selected reference set).
    Internal layers of the model explicitly mix information across the components of $X_t$, so that each decision can depend on the state of several solutions.
    The policy either produces a single action:
    $a_t \sim \pi_\theta(\cdot \mid X_t)$, or a joint action vector
    $A_t = (a_t^{(1)},\dots,a_t^{(P')}) \sim \Pi_\theta(\cdot \mid X_t)$ that involve several candidates.

\end{itemize}

The independent setting (Level~1) is simple and scalable: most existing single-trajectory NCO architectures, both constructive and improvement~\cite{bengio2021machine,kool2018attention,chen2019learning,garmendia2024marco}, already fall into this category and can be run in a population mode by treating $P$ solutions as a batch with little or no modification.
However, this regime misses key advantages of population methods: individuals do not exchange information, search effort cannot be adaptively reallocated across trajectories, and there is no explicit mechanism to maintain or exploit diversity in the population.
The richer Levels~2 and~3 are closer in spirit to classical population metaheuristics, but they introduce additional design challenges and, to the best of our knowledge, remain largely unexplored in NCO.
In what follows, we highlight three of the most important challenges.

\textbf{Population representation.}
As mentioned before, by a population we mean a set of candidate solutions that are considered together during the search.
This set can have variable size $P$, and different individuals may carry very different amounts of useful information (e.g., some are high-quality elites, others are diverse outliers).

Contextual and joint architectures (Levels 2 and 3) must therefore build an internal \emph{representation} of this population that:
(i) can handle variable population sizes $P$ without changing the model structure;
(ii) is invariant to the order in which individuals are listed (permutation symmetry), since the population is a set and not a sequence; and
(iii) preserves enough information about the population to support good subsequent decisions, while keeping memory and computation costs manageable.

At Level~2, the key challenge is to compress the whole population into per-individual context vectors $c_t^{(i)}$ that are both informative and cheap to compute.
The mapping $C_\theta(s_t^{(i)}, \mathcal{P}_t)$ in Eq.~\eqref{eq:level2-context} does exactly this: for each focal solution $s_t^{(i)}$, it looks at the population $\mathcal{P}_t$ and produces a short summary $c_t^{(i)}$ that encodes relevant information about the population from the perspective of solution $i$.
Designing $C_\theta$ means deciding which other solutions each individual can “see” (e.g., elites, diverse outliers, or $k$ nearest neighbors) and how to aggregate them into a fixed-size vector; if the summary is too coarse, the context is uninformative, whereas detailed summaries become expensive as problem or population size grows.

At Level~3, the population can be treated simply as a set (no explicit pairwise structure) or can be modeled with richer interactions between solutions. In the latter case, one must also decide how expressive these interactions should be (e.g., fully dense vs.\ sparse neighborhoods) while keeping the cost manageable. Ideally, the representation should be robust to permutations and varying population size $P$, but achieving this is non-trivial: it requires mapping a variable-size collection of variable-size solutions to a fixed number of embeddings without discarding too much information.

\textbf{Control structure and learning objective.}

Population-based search involves two kinds of decisions:
(i) \emph{local} decisions on individual solutions (e.g., which move to apply next), and
(ii) \emph{global} decisions about the population as a whole (e.g., which trajectories to restart, which solutions to keep, or how to distribute computation across individuals).

In a population-based RL setting, one could either define the objective separately for each individual solution or define a single objective for the population as a whole (e.g., based on the best or average solution quality).
In the latter case, a single global scalar objective aggregates the effects of many local and global decisions made by different individuals over time, which leads to a credit-assignment problem, i.e., it becomes unclear which specific decisions were responsible for a good or bad outcome.

At Level~2, control remains mostly local, but each decision is modulated by a context derived from the population.
Here, the challenge is to design learning signals that encourage the policy to exploit this population context, rather than ignoring it and reverting to purely single-solution behavior.

At Level~3, a single network can update many individuals based on their joint state, and the reward depends on the combined outcome of all of them.
In this regime, naïve global rewards provide very weak guidance, because each gradient step reflects many coordinated decisions at once.
A practical way to mitigate this could be to decompose the population update into a sequence of smaller decisions (for example, updating one individual or subset at a time), so that more informative intermediate feedback can be provided and credit can be assigned more precisely.

\textbf{Maintaining diversity under learning.}
A main motivation for using a population is to explore multiple regions of the search space in parallel. Yet, when a single neural policy is shared across all individuals and trained with a common RL objective that maximizes expected return, it tends to become confident and almost deterministic. In a population setting, this drives all individuals toward similar behaviors and the same basin of attraction, effectively wasting the population. Preventing this collapse requires explicit diversity mechanisms, acting either on the solutions (e.g., rewarding novelty or penalizing near-duplicates), on the behavior (e.g., assigning different roles or objectives to different individuals), or on the stochasticity of the policy (e.g., maintaining sufficient entropy).

In the next section, we address these challenges with concrete design choices that lead to a population-based neural architecture for combinatorial optimization.

\section{A Population-based Neural Architecture}
\label{sec:method}

\begin{figure*}[t]
\centering
\includegraphics[width=0.75\textwidth]{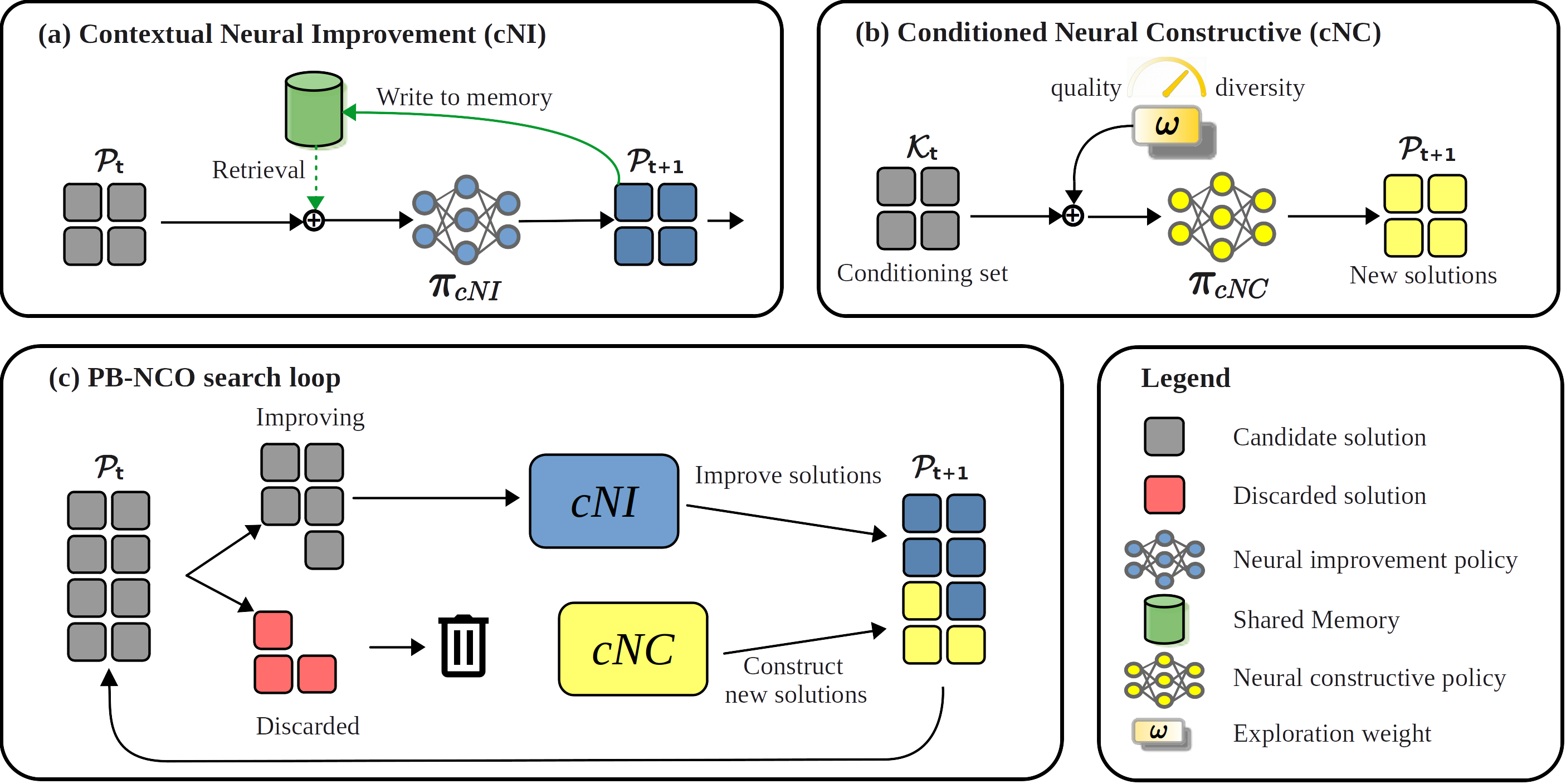}
\caption{\textbf{Overview of the proposed population-based NCO frameworks.}
\textbf{(a)} A contextual neural improvement policy (\textbf{cNI}) applies learned local moves to each solution, using information retrieved from a shared population memory of visited candidates.
\textbf{(b)} A conditioned neural constructive policy (\textbf{cNC}) samples new candidates given a conditioning set of solutions $\mathcal{K}$ and an exploration weight $\omega$, which controls the trade-off between solution quality and diversity.
\textbf{(c) PB-NCO} combines both components in a population loop, alternating cNI-driven improvement of existing solutions with cNC-driven restarts.
The cNI-improved individuals and the cNC-generated restart candidates jointly form the next population $\mathcal{P}_{t+1}$, which is used as input to the following iteration.}

\label{fig:p_nco_overview}
\end{figure*}

Building on the population awareness levels and the challenges outlined in Section~\ref{sec:design_challenges}, we propose two population-based components:
\begin{itemize}
    \item a \emph{contextual Neural Improvement} (\textbf{cNI}) method (Level~2), designed to \emph{improve} existing solutions by performing learned local search on each individual while reading and updating a central population memory; and
    \item a \emph{conditioned Neural Constructive} (\textbf{cNC}) method (Level~3), designed to \emph{generate} new solutions that are high-quality yet diverse with respect to a conditioning set of solutions.
\end{itemize}

The following subsections detail the two main components of our approach. We first present cNI, describing its interaction with the shared memory and its learning dynamics. We then introduce cNC, focusing on how it trades off solution quality and diversity, as well as its training scheme. Finally, we explain how these components are combined at inference time to form a more powerful, unified population-based framework, which we refer to as Population-Based Neural Combinatorial Optimization (\textbf{PB-NCO}).

\subsection{Contextual Neural Improvement Policy (Level~2)}
\label{subsec:impr}

In the improvement setting, the policy $\pi_{\text{cNI}}$\footnote{For notational convenience, we omit the parameter vector $\theta$.} conditions on both the instance and the current candidate solution $s_t^{(i)}$, and proposes a local modification (see Section~\ref{subsec:setup}). 
Applied independently to a batch of candidates, this yields a Level~1 method: $P$ isolated improvement trajectories with no information exchange. 
In contrast, we introduce a \emph{contextual} neural improvement policy (Level~2) that exploits a shared memory that provides a population-context.

Concretely, a central \emph{memory} $\mathcal{M}_t$ stores all solutions visited by the population. 
At each step, and for every solution, the policy retrieves from $\mathcal{M}_t$ a compact descriptor that summarizes the most relevant historical information for its current state. After using this information to propose a modification, the resulting solution is written back into memory (see Fig.~\ref{fig:p_nco_overview}a). This defines a simple read–write interaction pattern: individuals query the accumulated population history to guide their next move and, in turn, update that history for future queries.

This retrieval step is implemented as a \(k\)-nearest-neighbors (k-NN) query in solution space~\cite{garmendia2024marco}, defined by a problem-specific distance function \(d(\cdot,\cdot)\). Given the current solution \(s_t^{(i)}\), we select its \(k\) nearest neighbors \(\mathcal{N}_k(s_t^{(i)}) = \{ s^{(j)} \in \mathcal{M}_t : j \in \mathcal{I}_k \}\), where \(\mathcal{I}_k\) indexes the \(k\) entries in \(\mathcal{M}_t\) with smallest distance \(d(s_t^{(i)}, s^{(j)})\). These neighbors are aggregated into a fixed-size descriptor \(z_t^{(i)} = \sum_{j \in \mathcal{I}_k} \alpha_j s^{(j)}\), where the weights \(\alpha_j\) decay linearly with distance so that closer neighbors contribute more. Concretely, we compute distances \(d_j = d(s_t^{(i)}, s^{(j)})\), min--max normalize them to \(\tilde d_j = (d_j - d_{\min}) / (d_{\max} - d_{\min} + \varepsilon)\), invert them as \(w_j = 1 - \tilde d_j\) so that the closest neighbor attains weight \(w_j = 1\) and the farthest \(w_j = 0\), and finally normalize to \(\alpha_j = w_j / \sum_{\ell \in \mathcal{I}_k} w_\ell\).

The resulting descriptor, denoted \(z_t\) for brevity, is injected by augmenting the dynamic features of the current search state. Since \(z_t \in \mathbb{R}^{|\mathcal{V}|}\) is a solution-space vector, we use it as a \emph{per-node} channel. For node-based representations, we append \(z_{t,u}\) to node \(u\)'s features:
\(
\tilde h_u = [\,h_u \,;\, z_{t,u}\,]
\) for all nodes \(u \in \mathcal{V}\).
For edge-based representations, we instead concatenate an edge-wise scalar \(z_{t,uv}\) to the dynamic edge features, obtaining \(\tilde e_{uv} = [\,e_{uv} ; z_{t,uv}\,]\) for all edges \((u,v) \in \mathcal{E}\).

To ensure bounded memory use, the shared memory \(\mathcal{M}_t\) is maintained with a fixed capacity \(M_{\max}\). When this limit is reached, newly added solutions remove the oldest entries following a first-in–first-out policy.

To guide policy learning on top of this memory-augmented architecture, we define the per-step reward as the sum of two complementary components: an \emph{objective reward} that encourages genuine progress in solution quality and a \emph{repetition penalty} that discourages revisiting already explored solutions.

\paragraph{Objective reward} Rather than rewarding every small local gain as in Eq.~\ref{eq:ni_reward}, we use the best solution found in each trajectory as a moving reference.
Let
\begin{equation}
B_t^{(i)} = \max\{\,f(s_\tau^{(i)}) : 0 \le \tau \le t\,\}
\end{equation}
the best objective value achieved by individual $i$ since its most recent initialization or restart,
and let $s_{t+1}^{(i)}$ denote the post-action solution.  
The objective reward is then the non-negative improvement over this reference:
\begin{equation}
\label{eq:impr-obj}
R_{\mathrm{obj},t}^{(i)} = \max\{\,f(s_{t+1}^{(i)}) - B_t^{(i)},\, 0\,\}
\end{equation}
This choice makes the cumulative reward along a trajectory equal to the net improvement in $f$, which tolerates occasional worsening moves on the path to a better solution.

\paragraph{Repetition penalty}
To prevent cycling and increase exploration, we include a repetition penalty that activates when an individual revisits an already stored solution in the shared memory:
\begin{equation}
\label{eq:impr-rep}
R_{\mathrm{rep},t}^{(i)} \;=\;
\begin{cases}
-1, & \text{if } s_{t+1}^{(i)} \in \mathcal{M}_t,\\
\ \ 0, & \text{otherwise.}
\end{cases}
\end{equation}
We deliberately use a binary penalty rather than a distance- or diversity-based term. Explicitly maximizing diversity in a local improvement setting, would encourage moves that push trajectories far away from the visited solutions, potentially bypassing nearby improvements. In contrast, the binary penalty simply discourages revisiting previously seen solutions while allowing focused exploration in the surrounding neighborhood. In Section~\ref{subsec:rest}, we show that in a constructive setting, where solutions are generated from scratch, distance-based diversity penalties become more appropriate.

The total per-step reward is
\begin{equation}
\label{eq:impr-total}
R_{\mathrm{cNI}, t}^{(i)} \;=\; R_{\mathrm{obj},t}^{(i)} \;+\; w_{\mathrm{rep}}\,R_{\mathrm{rep},t}^{(i)}
\end{equation}
with a tunable weight $w_{\mathrm{rep}}\!\ge\!0$ that controls the strength of the repetition penalty.

\subsection{Conditioned Neural Constructive Policy (Level~3)}
\label{subsec:rest}

The second proposed component is a Level~3 method: a \emph{conditioned neural constructive} (cNC) policy designed to generate new candidates by trading off solution quality against distance  from a reference set $\mathcal{K}$ (see Fig.~\ref{fig:p_nco_overview}b). In population-based search, quality and diversity form an intrinsically bi-objective goal: pushing solely for quality causes the population to converge prematurely to a small set of nearly identical elites, while forcing diversity alone harms objective performance. To manage this, we draw inspiration from Conditioned Networks~\cite{abels2019dynamic}, originally proposed for multi-objective RL~\cite{felten2024multi}, and expose this trade-off directly to the network via explicit conditioning variables.

We instantiate this idea as a policy $\pi_{\text{cNC}}$ that extends the standard constructive architecture (Section~\ref{subsec:setup}) with two additional inputs: a \emph{conditioning set} $\mathcal{K}$ and an \emph{exploration weight} $\omega$.

At step $t$, the policy observes a reference set of solutions $\mathcal{K}_t = \{s^{(1)},\dots,s^{(|\mathcal{K}_t|)}\}$ (e.g., the current population or a subset of elites), which indicates which regions of the search space have already been explored. To encode this, we fix a maximum conditioning size $K_{\max}$ and augment the graph features with $K_{\max}$ additional channels. As a practical example, for node-based encodings, each solution $s^{(k)}$ is mapped to a binary indicator vector
\begin{equation}
    m_{u}^{(k)} \in \{0,1\}, \qquad u \in \mathcal{V},
\end{equation}
specifying whether node $u$ participates in $s^{(k)}$. The augmented node features become
\begin{equation}
    \tilde h_{u} 
\;=\;
\big[h_{u} \,;\, m_{u}^{(1)} \,;\, \dots \,;\, m_{u}^{(K_{\max})}\big],
\end{equation}
where unused channels are zero-padded if $|\mathcal{K}_t| < K_{\max}$.  
For edge-based representations, the same would be done for edge features.

To explicitly control this behavior, we introduce a scalar exploration weight $\omega \in [0,1]$, which is broadcast as an additional node feature:
\begin{equation}
    \tilde h_{u}
\;\leftarrow\;
[\,\tilde h_{u} \,;\, \omega\,], \qquad u\in\mathcal{V}.
\end{equation}
When $\omega \approx 0$, the policy is encouraged to exploit and produce high-quality solutions with minimal regard to $\mathcal{K}_t$. Conversely, when $\omega \approx 1$, it prioritizes diversity by generating solutions that are structurally distinct from the reference set. Intermediate values interpolate between these behaviors, encouraging solutions that balance quality and novelty.

\paragraph{Training Objective} 
To enable the policy to generalize across the full spectrum of the quality--diversity trade-off, we employ a randomized training scheme. For each problem instance in a training batch, we first generate a synthetic conditioning set $\mathcal{K}_t$ and sample a scalar exploration weight $\omega$. The policy $\pi_{\mathrm{cNC}}$ then constructs a candidate solution $s'$, conditioned on both the instance and the tuple $(\mathcal{K}_t,\omega)$.

Since the raw objective value and the solution distance can have different scales, we define the cNC reward using normalized quantities. Let $\tilde f_{\mathcal{I}}(s')$ denote the problem-normalized objective value for instance $\mathcal{I}$, and let $\tilde d(s',m)\in[0,1]$ denote the normalized distance between two solutions. In our experiments, $\tilde d$ is the normalized Hamming distance for both MC and MIS. The problem-specific definitions of $\tilde f_{\mathcal{I}}$ are given in Appendix~\ref{apendix_problem_implementation}.

The scalar reward used to train cNC is then
\begin{equation}
\label{eq:con_reward}
R_{\mathrm{cNC}}(s' \mid \mathcal{K}_t,\omega)
=
(1-\omega)\,\tilde f_{\mathcal{I}}(s')
+
\omega\,
\frac{1}{|\mathcal{K}_t|}
\sum_{m\in\mathcal{K}_t}
\tilde d(s',m).
\end{equation}
Here, $\omega=0$ emphasizes solution quality, $\omega=1$ emphasizes distance from the conditioning set, and intermediate values interpolate between the two normalized terms.

During training, we sample the exploration weight from a symmetric Beta distribution, $\omega \sim \mathrm{Beta}(\alpha, \beta)$ with $\alpha=\beta=0.2$. This distribution is bi-modal, concentrating probability mass near 0 and 1. This forces the network to master the distinct tasks of pure exploitation (quality maximization) and pure exploration (diversity maximization). For the policy-gradient update (Eq.~\ref{eq:pg}), the advantage \(A_t\) is computed using a baseline \(B(x_t)\) given by the mean reward of multiple candidate solutions \(s'\) generated for the same instance.

\paragraph{Inference and Integration} 
In this paper we use cNC in three complementary ways.
First, as a \emph{greedy constructive} heuristic: we set $\omega = 0$ and perform a single construction per instance, so that cNC behaves as a fast, quality-focused method.
Second, as a \emph{population-based generator}: we maintain an explicit population and, at each iteration, take the current population as the reference set $\mathcal{K}_t$. We then apply cNC to sample new candidates that will define the new population, while varying $\omega$ over time according to a defined schedule to interpolate between exploitation and exploration.
Third, cNC acts as a \emph{restart operator} within our full PB-NCO framework, as detailed in the following subsection.

\subsection{PB-NCO search scheme}
\label{sec_pb_nco_inf}

PB-NCO combines the contextual improvement policy (cNI) and the conditioned constructive policy (cNC) within a unified population-based search loop. As illustrated in Fig.~\ref{fig:p_nco_overview}c, each individual in the population repeatedly alternates between two phases: (i) \emph{local improvement} driven by cNI, and (ii) \emph{restart and diversification} guided by cNC.

During the improvement phase, cNI proposes local modifications to refine the current solution while using the shared memory to avoid immediate cycling. This mechanism steadily pushes individuals toward high-quality regions of the search space. However, improvement alone cannot ensure broad exploration: once an individual stops progressing, continuing to apply cNI yields little benefit.

To address this, PB-NCO introduces a learned restart mechanism. When an individual's trajectory becomes stagnant (detected via a patience counter) its current solution is discarded and replaced by a new candidate generated by cNC. The cNC receives a conditioning set $\mathcal{K}_t$, defined by the last $K_{\max}$ solutions inserted into the shared memory by the whole population, and an exploration weight $\omega$ to generate the new candidate.

\begin{algorithm}[tb]
\caption{Inference Pipeline}
\label{alg:inference}
\begin{algorithmic}[1]
\small
\STATE \textbf{Input:} instance $\mathcal{I}$; trained improvement policy $\pi_{\text{cNI}}$; trained conditioned constructor $\pi_{\text{cNC}}$; population size $P$; horizon $T_{\max}$; patience $N_{\text{pat}}$; initial exploration weight $\omega_{\text{start}}$; cooling factor $\phi$
\STATE \textbf{Initialize:}
\STATE $\mathcal{M}_0 \gets \varnothing$ \hfill\COMMENT{shared memory}
\STATE $s^{\star} \gets \varnothing$, \ $f(s^{\star}) \gets -\infty$ \hfill\COMMENT{best-so-far}
\FOR{$i \in \{1,\ldots,P\}$}
  \STATE $s^{(i)}_0 \gets \mathrm{Init}(\mathcal{I})$ \hfill\COMMENT{feasible initialization}
  \STATE $c^{(i)} \gets 0$ \hfill\COMMENT{consecutive non-improving steps}
  \STATE $B^{(i)}_0 \gets f(s^{(i)}_0)$ \hfill\COMMENT{per-individual best-so-far value}
  \STATE $\mathcal{M}_0 \gets \mathrm{UpdateMemory}(\mathcal{M}_0, s^{(i)}_0)$
\ENDFOR

\FOR{$t = 0$ \textbf{to} $T_{\max}-1$}
  \FOR{$i \in \{1,\ldots,P\}$}
    \IF{$c^{(i)} \ge N_{\text{pat}}$} 
      \STATE $\omega \gets \omega_{\text{start}}\!\left(1 - \frac{t}{T_{\max}}\right)^{\phi}$ \hfill\COMMENT{budget-aware schedule}
      \STATE $\mathcal{K}_t \gets \mathrm{SelectK}_{\mathrm{last}}(\mathcal{M}_t)$
      \STATE $s^{(i)}_{t+1} \gets \pi_{\text{cNC}}(\mathcal{I}, \mathcal{K}_t, \omega)$ \hfill\COMMENT{construct new solution}
      %\STATE $c^{(i)} \gets 0$, \ $B^{(i)}_{t+1} \gets \max\{B^{(i)}_t,\, f(s^{(i)}_{t+1})\}$
      \STATE $c^{(i)} \gets 0$, \ $B^{(i)}_{t+1} \gets f(s^{(i)}_{t+1})$ \hfill\COMMENT{restart counter and best}
    \ELSE
        \STATE $s^{(i)}_{t+1} \gets \pi_{\text{cNI}}(\mathcal{I}, s^{(i)}_t, \mathcal{M}_t)$ \hfill\COMMENT{perform local move}
        \IF{$f(s^{(i)}_{t+1}) > B^{(i)}_t$} 
          \STATE $c^{(i)} \gets 0$, \ $B^{(i)}_{t+1} \gets f(s^{(i)}_{t+1})$ 
        \ELSE
          \STATE $c^{(i)} \gets c^{(i)} + 1$, \ $B^{(i)}_{t+1} \gets B^{(i)}_t$
        \ENDIF
    \ENDIF
    \STATE $\mathcal{M}_{t+1} \gets \mathrm{UpdateMemory}(\mathcal{M}_t, s^{(i)}_{t+1})$
    \IF{$f(s^{(i)}_{t+1}) > f(s^{\star})$}
      \STATE $s^{\star} \gets s^{(i)}_{t+1}$ \hfill\COMMENT{update best-so-far}
    \ENDIF
  \ENDFOR
\ENDFOR
\STATE \textbf{Output:} $s^{\star}$
\end{algorithmic}
\end{algorithm}

Algorithm~\ref{alg:inference} summarizes the full inference pipeline. The algorithm takes as input an instance $\mathcal{I}$, trained policies $\pi_{\text{cNI}}$ and $\pi_{\text{cNC}}$, and several runtime parameters: population size $P$, iteration budget $T_{\max}$, patience threshold $N_{\text{pat}}$, and an exploration-weight schedule specified by $\omega_{\text{start}}$ and cooling exponent $\phi$. PB-NCO begins by generating $P$ feasible starting solutions, either randomly or using a highly exploitative call to cNC, and inserting them into the shared memory.

The main loop (lines 12–32) then iterates over all individuals. As long as an individual continues to improve, cNI generates local moves conditioned on the memory. When no improvement has been observed for $N_{\text{pat}}$ consecutive steps, the individual is restarted: a conditioning set $\mathcal{K}_t$ with the last $K_{\max}$ solutions is selected, a budget-aware exploration weight $\omega$ is computed, and cNC constructs a new solution conditioned on $(\mathcal{I}, \mathcal{K}_t, \omega)$. Each new solution, whether obtained through improvement or restart, is written to the shared memory, and the global best solution $s^\star$ is updated whenever a better objective value is encountered. The algorithm outputs this best solution at termination.

\section{Benchmark Problems}
\label{sec:benchmark_problems}

We consider two graph-based combinatorial optimization problems within the setup of Section~\ref{subsec:CO_setup}. 
In both problems, an instance is \(\mathcal{I}\equiv G=(\mathcal{V},\mathcal{E})\) (optionally with edge weights \(w_{uv}\!>\!0\)). 
Solutions are binary vectors \(\mathbf{s}\in\{0,1\}^{|\mathcal{V}|}\), and we maximize \(f_{\mathcal{I}}(\mathbf{s})\) over a feasible set \(\mathcal{S}(\mathcal{I})\).

\textbf{Maximum Cut (MC)} seeks a binary labeling of nodes that places the maximum number of adjacent nodes on opposite sides; edges “cut” by the partition contribute to the score.

The feasible set is all assignments, \(\mathcal{S}_{\mathrm{MC}}(\mathcal{I})=\{0,1\}^{|\mathcal{V}|}\). 
The objective function weights the edges crossing the bipartition:
\begin{equation}
    f_{\mathcal{I}}(\mathbf{s})
    \;=\;
    \sum_{(u,v)\in \mathcal{E}} w_{uv}\,\mathbf{1}\{s_u \neq s_v\},
    \qquad
    \mathbf{s}\in \mathcal{S}_{\mathrm{MC}}(\mathcal{I})
\end{equation}
where $\mathbf{1}\{\cdot\}$ denotes the indicator function. Throughout the experiments we consider unweighted instances and set \(w_{uv}=1\) for all \((u,v)\in \mathcal{E}\).

\textbf{Maximum Independent Set (MIS)} selects the largest subset of mutually non-adjacent nodes, so no chosen pair may share an edge. The difficulty lies in respecting feasibility while growing the set.

The feasible set enforces independence:
\begin{equation}
\mathcal{S}_{\mathrm{MIS}}(\mathcal{I})
=
\big\{\mathbf{s}\in\{0,1\}^{|\mathcal{V}|}:\ \mathbf{s}_u+\mathbf{s}_v \le 1 \ \ \forall (u,v)\in \mathcal{E}\big\},
\end{equation}
and the objective is to maximize the set size:
\begin{equation}
f_{\mathcal{I}}(\mathbf{s})
\;=\;
\sum_{u\in \mathcal{V}}\mathbf{s}_u,
\qquad
\mathbf{s}\in \mathcal{S}_{\mathrm{MIS}}(\mathcal{I})
\end{equation}

Appendix~\ref{apendix_problem_implementation} provides problem-specific implementation details for MC and MIS, including decoding procedures for cNI/cNC, reward normalization schemes, diversity metrics, and the exact feature encodings used in our experiments.

\section{Experimental Protocol}
\label{sec:exp_protocol}

This section describes the experimental protocol used to evaluate the proposed population-based NCO framework. We specify the evaluated components, training distributions, evaluation instances, inference settings, baselines, hardware, and reporting metrics.

\textbf{Evaluated components.}
We evaluate cNI as a standalone improvement method, cNC both as a single-pass greedy constructive heuristic (cNC$_{\text{greedy}}$) and as a population-based generator (cNC$_{\text{pop}}$), and PB-NCO as the full framework described in Section~\ref{sec_pb_nco_inf}.

\textbf{Neural Network Architecture.} We use the \textbf{Graph Transformer} (GT)~\cite{dwivedi2020generalization} as the backbone encoder for both the cNI and cNC components, with lightweight multi-layer perceptrons as task-specific decoders.
The full list of architectural hyperparameters are shown in Appendix~\ref{sec:appendix_hyperparameters}.

\textbf{Training distributions and population setup.}
For the improvement component (cNI), we train a single policy on randomly generated Erdős--Rényi (ER) graphs with edge probability $0.15$ and sizes ranging from 50 to 300 nodes.
This same cNI policy is used in all ER and RB evaluations, so its RB results test transfer across both graph size and graph family.
For the constructive component (cNC), we train two separate policies: one on the same ER distribution, and another on RB graphs with 200–300 nodes~\cite{xu2000exact}. cNC is evaluated with the family-matched policy: the ER-trained cNC is used on ER graphs, and the RB-trained cNC is used on RB graphs.

\textbf{Evaluation protocol.}
At evaluation time, all models are applied to larger graphs without any additional training.
More concretely, and following recent studies~\cite{ahn2020learning,bother2021s,zhang2023let}, we assess generalization on larger ER graphs (a dataset of 128 instances with 700–800 nodes) and on more challenging RB graphs~\cite{xu2000exact} (a dataset of 500 instances with 800–1200 nodes).

For PB-NCO inference, we set a patience threshold of $N_{\text{pat}} = 500$ non-improving steps before triggering a restart, and use a linear cooling schedule for the cNC exploration weight, decreasing $\omega$ from $1.0$ to $0.0$ over the allocated budget.
By default, populations are initialized from random feasible solutions; an alternative initialization using cNC is analyzed separately.

\textbf{Baselines.}
For a more comprehensive comparison, we include exact methods using the GUROBI solver~\cite{gurobi}, simple greedy heuristics that iteratively add the item with the largest marginal gain, metaheuristics such as Genetic Algorithms (GA)~\cite{kramer2017genetic} and Particle Swarm Optimization (PSO)~\cite{kennedy1995particle}, specialized algorithms like BURER~\cite{burer2002rank} for MC and K$_A$MIS~\cite{lamm2016finding} for MIS, and learning-based methods including S2V-DQN~\cite{khalil2017learning}, ECO-DQN~\cite{barrett2020exploratory}, FlowNet~\cite{zhang2023let}, ANYCSP~\cite{tonshoff2023one}, MARCO~\cite{garmendia2024marco} for MC, and additionally DGL~\cite{bother2021s}, LwD~\cite{ahn2020learning}, INTEL~\cite{li2018combinatorial} and DiffUCO~\cite{sanokowski2024diffusion} for MIS. Together, these methods span a broad spectrum of algorithms, including the state-of-the-art techniques.

\textbf{Hardware.}
Exact methods, heuristics, and metaheuristics have been executed using a cluster with 32 \textit{Intel Xeon X5650} CPUs.
Learning-based methods have been implemented using \textit{PyTorch 2.0}, and a \textit{Nvidia A100} GPU has been used to train the models and perform inference.

\textbf{Code availability.}
The source code and implementation of the proposed methods are publicly
available at \url{https://github.com/TheLeprechaun25/PB-NCO}.

\section{Results}
\label{sec:results}

This section presents the empirical results. 
We first report fixed-budget and anytime performance against classical, specialized, and neural baselines. 
We then analyze component ablations, search dynamics through diversity and cycling metrics, hyperparameter sensitivity, computational overhead, and the quality--diversity behavior of cNC.

\subsection{Effectiveness Experiments}

This section reports performance results for the proposed neural population methods on \textbf{MC} and \textbf{MIS}, comparing them against the baselines described in Section~\ref{sec:exp_protocol}.
For anytime methods, which can keep improving a solution as more computation is allocated, we impose a common \emph{1-minute per-instance} budget to obtain compute-matched comparisons.
For single-pass constructive methods, including both heuristic and learned NC approaches, we report the actual time required to generate one solution, since they do not naturally exploit the full budget.
We further provide \emph{anytime} curves, plotting the best-so-far objective against wall-clock time; for these curves, each learned baseline is run with the iteration limit recommended in its original paper or implementation.

\renewcommand{\arraystretch}{1.02}
\setlength{\tabcolsep}{.25em}
\begin{table}[!tbh]
\caption{\textbf{Performance on Maximum Cut (MC) and Maximum Independent Set (MIS).}
We report average objective value, ratio, and runtime under a 1-minute budget for anytime method on ER graphs with 700--800 nodes and RB graphs with 800--1200 nodes.
*Ratios are computed with respect to the best non-proposed method.
The best result in each block is highlighted in bold.
$^\dagger$GUROBI is run as a time-limited exact solver and is not assumed to be optimal.}
\vspace{0.1cm}
\centering 
\scriptsize
\begin{tabular}{@{}c l l r r r r r r@{}}
\toprule
 & & & \multicolumn{3}{c}{\textbf{ER700-800}} & \multicolumn{3}{c}{\textbf{RB800-1200}} \\
\cmidrule(lr){4-6}\cmidrule(lr){7-9}
\textbf{} & \textbf{Method} & \textbf{Type} & 
\textbf{Obj.} & \textbf{Ratio} & \textbf{Time} &
\textbf{Obj.}  & \textbf{Ratio} & \textbf{Time} \\
\midrule
\multirow{12}{*}{\rotatebox{90}{\textbf{Maximum Cut (MC)}}}
 & GUROBI & Exact$^\dagger$ 
   & 23420.17 & 0.966 & 1.00m 
   & 20290.08 & 0.643 & 1.00m \\
 & Greedy & Heuristic
   & 23774.79 & 0.980 & 0.03 s 
   & 30619.32 & 0.971 & 0.04 s \\
 & GA & Metaheuristic
   & 24211.64 & 0.999 & 1.00m 
   & 31542.89 & 1.000 & 1.00m \\
 & PSO & Metaheuristic
   & 24201.78 & 0.999 & 1.00m 
   & *31544.80 & 1.000 & 1.00m \\
 & BURER & Specialized
   & *24235.93 & 1.000 & 1.00m
   & 29791.52 & 0.944 & 1.00m \\
\cmidrule(lr){2-9}
 & S2V-DQN & RL / NC
   & 21581.79 & 0.890 & 10.1 s 
   & 22014.93 & 0.698 & 13.4 s \\
 & FlowNet & UL / NC
   & 23466.09 & 0.968 & 1.00m
   & 26554.93 & 0.842 & 1.00m \\
 & ECO-DQN & RL / NI
   & 23777.13 & 0.981 & 1.00m 
   & 29103.38 & 0.923 & 1.00m \\
 & ANYCSP & RL / NI
   & 23837.95 & 0.984 & 1.00m
   & 31466.11& 0.998 & 1.00m \\
 & MARCO & RL / NI
   & 24176.62 & 0.998 & 1.00m 
   & 30790.39 & 0.976 & 1.00m \\
   \cmidrule(lr){2-9}
 & cNC$_{\text{greedy}}$  & RL / NC
   & 23968.41 & 0.989 & 0.01 s
   & 31016.34 & 0.983 & 0.02 s \\
 & cNC$_{\text{pop}}$  & RL / NC
   & 24225.75 & 1.000 & 1.00m
   & 31545.60 & 1.000 & 1.00m \\
 & cNI  & RL / NI
   & 24221.72 & 0.999 & 1.00m
   & 31544.85 & 1.000 & 1.00m \\
 & PB-NCO  & RL / NI+NC
   & \textbf{24266.33} & \textbf{1.001} & 1.00m
   & \textbf{31547.50} & \textbf{1.000} & 1.00m \\
\midrule
\midrule
\multirow{14}{*}{\rotatebox{90}{\textbf{Maximum Independent Set (MIS)}}}
& GUROBI & Exact$^\dagger$ 
   & 43.47 & 0.966 & 1.00m 
   & 40.90 & 0.947 & 1.00m \\
 & Greedy & Heuristic
   & 38.85 & 0.863 & 0.05 s 
   & 37.78 & 0.875 & 0.06 s \\
 & GA & Metaheuristic  
   & 43.97 & 0.977 & 1.00m 
   & 41.39 & 0.959 & 1.00m \\
 & PSO & Metaheuristic
   & 43.69 & 0.971 & 1.00m 
   & 41.19 & 0.955 & 1.00m \\
 & K{\tiny A}MIS & Specialized  
   & *44.98 & 1.000 & 1.00m 
   & *\textbf{43.15} & \textbf{1.000} & 1.00m \\
\cmidrule(lr){2-9}
 & DGL & SL / Hybrid 
   & 38.71 & 0.861 & 11.0 s 
   & 32.32 & 0.750 & 2.61 s \\
 & INTEL & SL / Hybrid
   & 41.13 & 0.913 & 10.0 s 
   & 34.24 & 0.794 & 2.44 s \\
 & LwD & RL / NC  
   & 41.17 & 0.915 & 4.00 s 
   & 34.50 & 0.799 & 0.86 s \\
 & FlowNet & UL / NC 
   & 42.19 & 0.938 & 1.00m 
   & 38.30 & 0.888 & 1.00m \\
 & DiffUCO & UL / NC 
   & 43.40 & 0.965 & 1.00m 
   & 40.16 & 0.931 & 1.00m \\
 & MARCO & RL / NI  
   & 44.73 & 0.994 & 1.00m
   & 40.47 & 0.938 & 1.00m \\
\cmidrule(lr){2-9}
 & cNC$_{\text{greedy}}$  & RL / NC
   & 38.96 & 0.866 & 1.70 s 
   & 34.37 & 0.797 & 1.40 s \\
 & cNC$_{\text{pop}}$  & RL / NC
   & 39.64 & 0.881 & 1.00m 
   & 37.37 & 0.866 & 1.00m \\
 & cNI  & RL / NI
   & 44.99 & 1.000 & 1.00m
   & 40.28 & 0.933 & 1.00m \\
 & PB-NCO  & RL / NI+NC
   & \textbf{45.38} & \textbf{1.009} & 1.00m
   & 40.97 & 0.949 & 1.00m \\
\bottomrule
\end{tabular}
\label{table:mc_mis_perf}
\end{table}

Table~\ref{table:mc_mis_perf} reports results on ER700-800 and RB800-1200 instances.
Each cell shows the mean objective, the ratio to the reference objective (given by the best baseline marked with *), and the corresponding runtime.
For \textbf{MC}, PB-NCO obtains the best average objective on both datasets.
On ER700--800, PB-NCO achieves $24266.33 \pm 1.48$, compared with $24235.93$ for BURER, and reaches the best or tied-best value on $83\%$ of instances, compared with $38\%$ for BURER.
On RB800--1200, PB-NCO achieves $31547.50 \pm 2.32$, compared with $31544.80$ for PSO, and reaches the best or tied-best value on $72\%$ of instances, compared with $32\%$ for PSO.
Thus, the ER gain is clear, while the RB gain is small and should be interpreted as a marginal average improvement.
For \textbf{MIS}, PB-NCO achieves the best overall score on ER and remains competitive on RB, where K\textsubscript{A}MIS remains stronger.

Regarding the individual components, cNC in its greedy form (cNC\textsubscript{greedy}, $\omega{=}0$, argmax) serves as an extremely fast and reasonably strong constructive heuristic. 
In population mode (cNC\textsubscript{pop}, sampling), cNC further benefits from repeated sampling and diversity-aware conditioning, although these results use the family-matched cNC policy for each graph family. 
By contrast, cNI uses the same ER-trained policy in both ER and RB evaluations, testing transfer across size and, for RB, graph family. 
Both components can match or surpass most learning-based competitors and classical metaheuristics, and their combination in PB-NCO gives the strongest overall performance.

\begin{figure*}
\centering
% top row
\subfloat[MC — ER700–800]{%
  \includegraphics[width=0.48\textwidth]{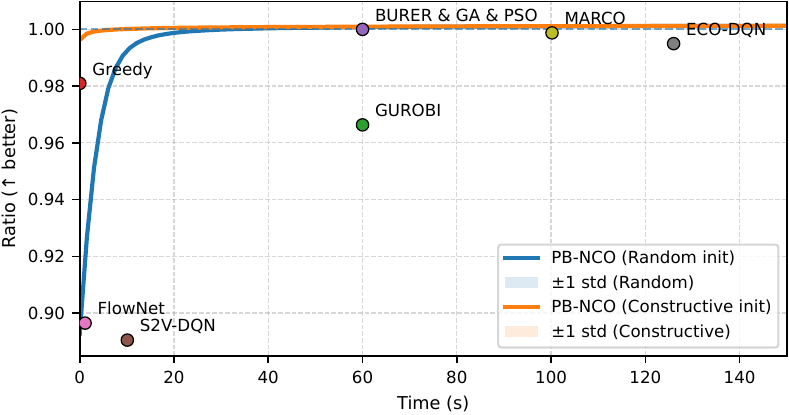}%
  \label{fig:anytime_mc_er}
}\hfill
\subfloat[MC — RB800–1200]{%
  \includegraphics[width=0.48\textwidth]{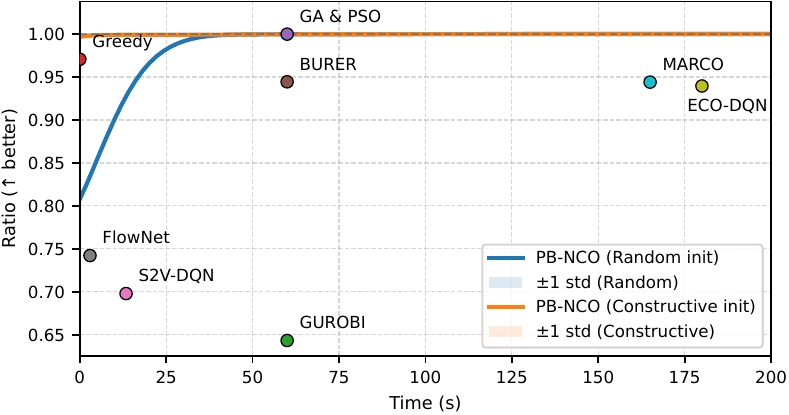}%
  \label{fig:anytime_mc_rb}
}\\[0.05ex]
% bottom row
\subfloat[MIS — ER700–800]{%
  \includegraphics[width=0.48\textwidth]{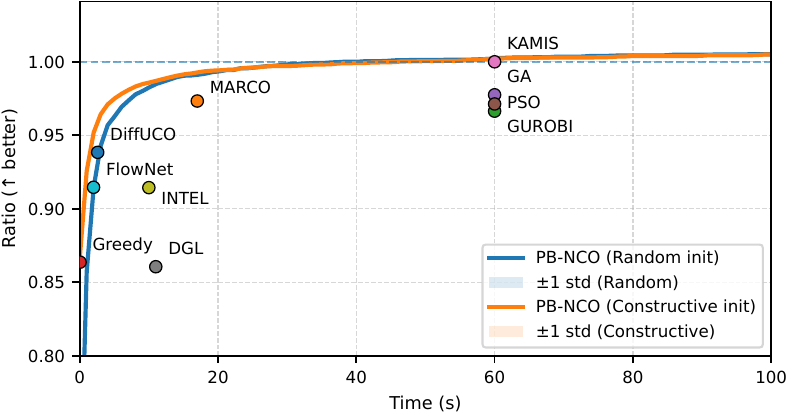}%
  \label{fig:anytime_mis_er}
}\hfill
\subfloat[MIS — RB800–1200]{%
  \includegraphics[width=0.48\textwidth]{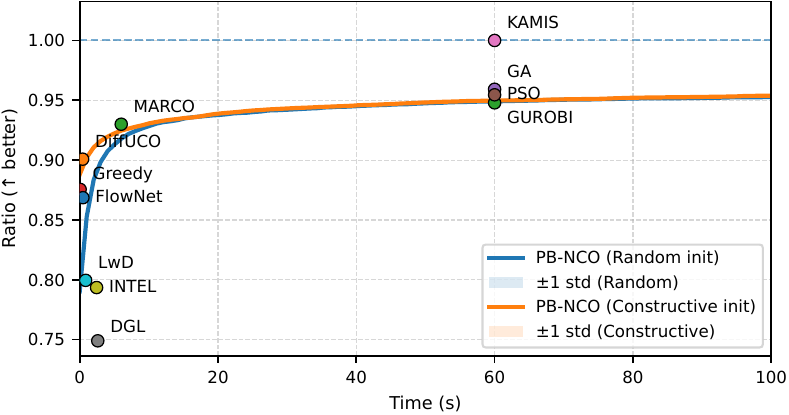}%
  \label{fig:anytime_mis_rb}
}
\caption{\textbf{Anytime performance against baselines.} We report the average best-so-far ratio over time for Maximum Cut (MC) and Maximum Independent Set (MIS) on ER graphs with 700--800 nodes and RB graphs with 800--1200 nodes. The ratio is computed with the reference objective used in Table~\ref{table:mc_mis_perf}. 
Solid lines show the mean over five runs, and shaded bands show one standard deviation; the bands are narrow because variability across runs is small. 
Markers denote baseline methods, with labels placed next to each marker for readability.
}

\label{fig:anytime_performance_baselines}
\end{figure*}

To provide a more fine-grained comparison across time scales, Fig.~\ref{fig:anytime_performance_baselines} shows anytime performance (best-so-far objective ratio) for PB-NCO under two initializations (Random and Constructive) together with the strongest baselines.
The constructive initialization uses cNC to generate an initial population in an iterative way, treating previously built solutions as visited and increasing the exploration weight linearly from $\omega{=}0$ to $\omega{=}0.5$.

Across all MC datasets and on MIS–ER, PB-NCO delivers consistently superior solution quality and maintains strong anytime behavior relative to both single-shot and iterative competitors.
Constructive initialization yields better early-time performance and reaches its plateau slightly earlier, whereas random initialization starts weaker but catches up over longer budgets; by the end of the run both variants are nearly indistinguishable.
Variability is very low overall and slightly smaller for random initialization (ratio standard deviation of order $10^{-4}$ at $60\,\mathrm{s}$ in MC), indicating stable behavior across runs.

\begin{figure*}
\centering
% top row
\subfloat[MC — ER700–800]{%
  \includegraphics[width=0.48\textwidth]{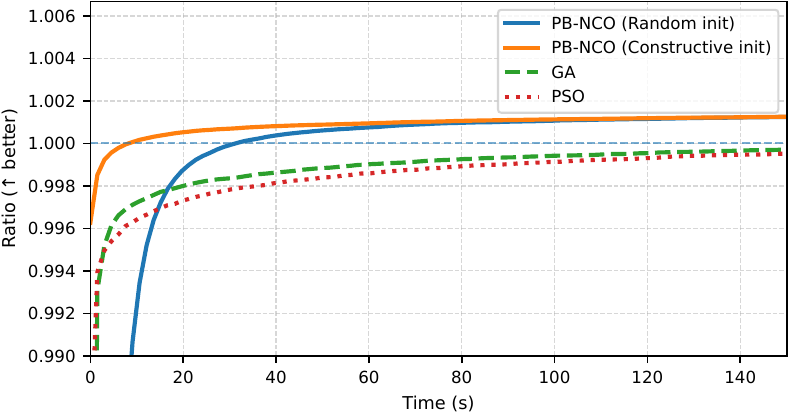}%
  \label{fig:anytime_ga_pso_mc_er}
}\hfill
\subfloat[MIS — ER700–800]{%
  \includegraphics[width=0.48\textwidth]{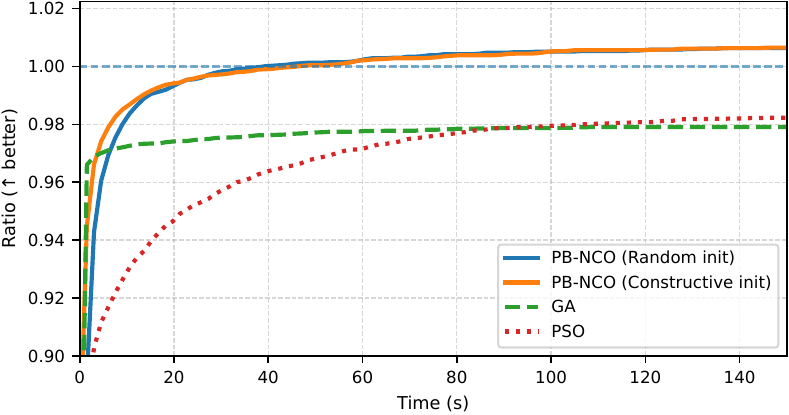}%
  \label{fig:anytime_ga_pso_mis_er}
}\hfill
\caption{\textbf{Anytime performance against metaheuristics.} 
Solid lines denote the average best-so-far ratio (current objective / reference objective) over five runs; shaded bands denote the standard deviation. Results are shown for both MC and MIS in ER700-800 graphs.
}
\label{fig:anytime_performance_mh}
\end{figure*}

Figure~\ref{fig:anytime_performance_mh} contrasts PB-NCO with population-based metaheuristics (GA and PSO) using the same population size, operators, and time budget.
PB-NCO converges faster and reaches higher final quality in these ER experiments, suggesting that the learned population mechanisms can provide advantages over classical population updates under the same budget.

\subsection{Main Ablation Study}

In this section we quantify the contribution of PB-NCO's core components via controlled ablations.
Unless stated otherwise, all variants use the same training protocol, seeds, population size ($P=20$), and a 5-minute budget.
We consider three variants:
\textbf{(i) Level 1 + Memory} (\emph{L.1+Mem.}), where individuals use private memories and do not exchange information through the shared memory, isolating the effect of population-level communication;
\textbf{(ii) cNI}, which uses the shared-memory improvement policy but disables restarts, isolating the effect of cNC-based diversification;
and \textbf{(iii) Random-Restarts} (\emph{Rand.-Rest.}), which keeps the cNI policy and restart schedule but replaces cNC restarts with random feasible solutions, isolating the value of learned constructive restarts.

\renewcommand{\arraystretch}{1.02}
\setlength{\tabcolsep}{.28em}
\begin{table}[!tbh]
\caption{\textbf{Ablation study on Maximum Cut (MC) and Maximum Independent Set (MIS).}
Performance of PB-NCO under different component removals and configuration changes.
All variants are evaluated under the same 5-minute budget.
Objective values are reported as mean $\pm$ standard deviation over five runs; ratios are computed from the mean objective.}
\vspace{0.1cm}
\centering 
\scriptsize
\begin{tabular}{@{}c l c r c r@{}}
\toprule
 & & \multicolumn{2}{c}{\textbf{ER700-800}} & \multicolumn{2}{c}{\textbf{RB800-1200}} \\
\cmidrule(lr){3-4}\cmidrule(lr){5-6}
\textbf{} & \textbf{Method} & 
\textbf{Obj.} $\uparrow$ & \textbf{Ratio} $\uparrow$ &
\textbf{Obj.} $\uparrow$ & \textbf{Ratio} $\uparrow$ \\
\midrule
\multirow{4}{*}{\rotatebox{90}{\textbf{MC}}}
 & L.1+Mem. 
   & $24230.26 \pm 2.87$ & 1.000
   & $31545.68 \pm 2.41$ & 1.000 \\
 & cNI 
   & $24232.17 \pm 1.74$ & 1.000
   & $31545.97 \pm 1.91$ & 1.000 \\
 & Rand.-Rest. 
   & $24269.50 \pm 1.62$ & 1.001
   & $31547.60 \pm 2.39$ & 1.000 \\
 & PB-NCO  
   & $\mathbf{24275.17 \pm 1.24}$ & \textbf{1.002}
   & $\mathbf{31548.03 \pm 2.11}$ & \textbf{1.000} \\
\midrule
\midrule
\multirow{4}{*}{\rotatebox{90}{\textbf{MIS}}}
 & L.1+Mem. 
   & $45.12 \pm 0.04$ & 1.003
   & $40.40 \pm 0.04$ & 0.936 \\
 & cNI 
   & $45.19 \pm 0.03$ & 1.004
   & $40.67 \pm 0.02$ & 0.942 \\
 & Rand.-Rest. 
   & $45.29 \pm 0.05$ & 1.006
   & $40.75 \pm 0.04$ & 0.944 \\
 & PB-NCO  
   & $\mathbf{45.44 \pm 0.02}$ & \textbf{1.010}
   & $\mathbf{41.44 \pm 0.01}$ & \textbf{0.960} \\

\bottomrule
\end{tabular}
\label{table:mc_mis_perf_ablations}
\end{table}

Table~\ref{table:mc_mis_perf_ablations} summarizes the ablation results, reporting average objective value, its standard deviation, and the average ratio to the strongest baseline for each dataset.
Across problems and datasets, all three variants underperform the full PB-NCO, confirming that shared memory, learned constructive restarts, and the restart scheduler each contribute to performance.

A notable observation concerns the per-individual mean objective. The \textit{L1+Mem.} variant exhibits a higher average across the $20$ individuals than PB-NCO, yet its elite (best-of-population) solution is consistently worse. 
This is expected: PB-NCO deliberately spreads the population via memory-conditioned coordination and restarts, allocating some individuals to intensify around promising regions while others probe diverse areas. 
Interestingly, we have seen that the resulting higher inter-individual variance lowers the average objective value of the individuals but raises the maximum, precisely the quantity that matters in population-based search.

\subsection{Search Dynamics: Diversity and Cycling}

This section analyzes two search-dynamics mechanisms behind PB-NCO: how cNC controls the diversity of generated solutions, and how the cNI repetition penalty reduces cycling.

\begin{figure}[!t]
\centering
\includegraphics[width=0.48\textwidth]{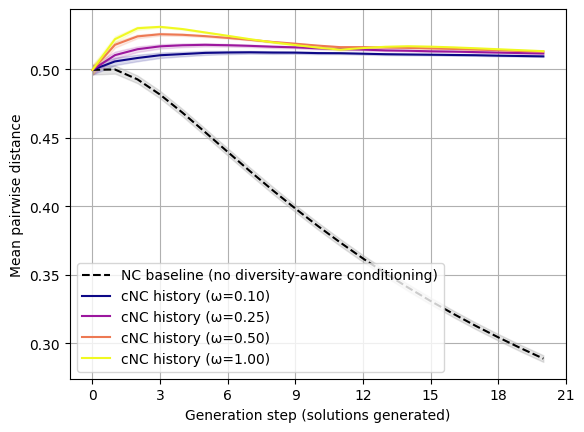}
\caption{\textbf{Diversity during cNC search.}
Mean pairwise normalized Hamming distance over all solutions visited so far, for the NC baseline (no diversity-aware conditioning) and for cNC with several exploration weights $\omega$.}
\label{fig:diversity_trajectory}
\end{figure}

\textbf{Diversity in cNC.}
Figure~\ref{fig:diversity_trajectory} compares the diversity of solutions generated by cNC with different exploration weights $\omega$ and by an unconditioned NC baseline.
All methods start from the same random initial population, after which we generate one solution at a time and measure the mean pairwise normalized Hamming distance over the visited solutions.
The unconditioned baseline rapidly collapses to a narrower region of the search space, while cNC maintains substantially higher diversity across the run.
This shows that conditioning on $\omega$ provides an effective mechanism for controlling exploration during constructive search.

\begin{table}[t]
\centering
\scriptsize
\setlength{\tabcolsep}{4pt}
\renewcommand{\arraystretch}{1.05}
\caption{\textbf{Effect of the cNI repetition penalty.}
Ratio is computed relative to the best objective among the tested $w_{\rm rep}$ values within each regime.
Revisit freq. is the exact-revisit frequency over a five-minute run.}
\label{tab:wrep_ablation}
\begin{tabular}{@{}lcc cc cc@{}}
\toprule
& \multicolumn{2}{c}{\textbf{Train-low ER50}} 
& \multicolumn{2}{c}{\textbf{Train-high ER300}} 
& \multicolumn{2}{c}{\textbf{Eval ER700--800}} \\
\cmidrule(lr){2-3}
\cmidrule(lr){4-5}
\cmidrule(lr){6-7}
$w_{\rm rep}$ 
& \textbf{Ratio} 
& \textbf{Revisit freq.}
& \textbf{Ratio} 
& \textbf{Revisit freq.}
& \textbf{Ratio} 
& \textbf{Revisit freq.} \\
\midrule
$0.0$ & 1.000 & 0.919 & 0.998 & 0.027 & 0.981 & 0.000 \\
$0.1$ & 1.000 & 0.872 & 0.999 & 0.004 & 0.998 & 0.000 \\
$1.0$ & 1.000 & 0.318 & 1.000 & 0.000 & 1.000 & 0.000 \\
$5.0$ & 1.000 & 0.139 & 1.000 & 0.000 & 0.997 & 0.000 \\
\bottomrule
\end{tabular}
\end{table}

\textbf{Repetition in cNI.} Here we measure the effect of training the cNI with different 
repetition weights $w_{\rm rep}$.
Table~\ref{tab:wrep_ablation} shows that removing the repetition penalty causes frequent revisits on training-size graphs, especially on ER50.
Larger penalties strongly reduce revisiting without degrading quality in the training regimes.
On ER700--800, revisits are already rare, but nonzero penalties still improve performance, suggesting that the penalty also encourages more useful search behavior beyond simply avoiding repeated states.

\subsection{Hyperparameter Sensitivity Analysis}

Beyond the main ablations, we examine sensitivity to two hyperparameters that affect performance: population size, and restart patience (the number of non-improving steps before triggering a restart).

\begin{figure}[!t]
\centering
\includegraphics[width=0.48\textwidth]{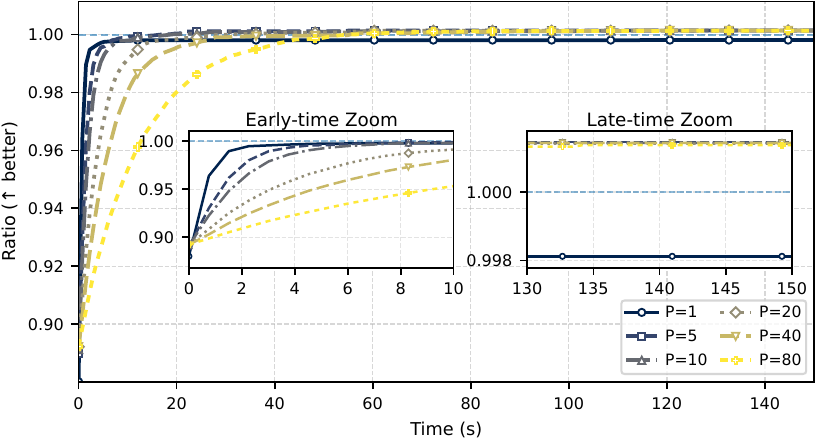}
\caption{\textbf{Effect of population size.}
Average best-so-far ratio over time, comparing different population sizes $P$.}
\label{fig:pop_size_perf}
\end{figure}

\textbf{Population Size.} We present the results of various population sizes under the same time budget in Fig. \ref{fig:pop_size_perf}. 
As expected, smaller populations improve faster at the beginning because, under a fixed time budget, the algorithm completes more iterations of improvement per candidate, so good solutions appear sooner. 
After the initial improvement phase, the curves begin to flatten: in our experiments, populations of $P\ge5$ achieve very similar final performance, suggesting diminishing returns beyond a small but non-trivial population. In contrast, the run with $P=1$, which cannot exploit population-level mechanisms, is consistently worse at the end of the execution.

\textbf{Restarting Patience.} We study the restart patience $N_{\text{pat}}$, the number of consecutive non-improving iterations an individual is allowed before restart, on ER700–800 with $N_{\text{pat}}\in\{100,500,1000,5000\}$. A low value can cause premature exploration, potentially missing out on good solutions, while a high value can result in excessive computation within a local region of the search space without yielding significant improvements.

In our runs, moderate settings performed best: for MC–ER, $N_{\text{pat}}{=}500$ and $1000$ produced nearly identical top results ($\approx 24266$), whereas both $100$ and $5000$ were consistently lower ($\approx 24254$ and $\approx 24256$). 
MIS–ER showed the same ordering ($\approx 45.38/45.37$ at $500/1000$ vs.\ $\approx 45.11/45.19$ at $100/5000$). 
Based on these observations, setting the patience to the number of nodes in the graph instance ($N_{\text{pat}}\approx |\mathcal{V}|$) serves as a simple yet effective rule of thumb.

\subsection{Computational Overhead and Memory Scalability}
\label{subsec:overhead_memory}

PB-NCO adds two overheads with respect to a standard neural method: storing visited solutions and retrieving nearest neighbors from memory. For binary graph problems, storing $M_t$ solutions costs $O(M_t|\mathcal{V}|)$ memory, and exact $k$NN retrieval with normalized Hamming distance costs $O(PM_t|\mathcal{V}|)$ per population iteration before batching, where $P$ is the population size.

We measured the time required to process a fixed budget of $10|\mathcal{V}|$ population iterations on ER700--800 graphs. Runtime scaled almost linearly with $P$, with fit $t(P)=a+bP$, where $a\approx16.2$s, $b\approx10.95$s per individual, and $R^2\approx0.99991$. The shared memory grows by one entry per inserted solution; thus, after $T$ population iterations, $M_{\mathrm{peak}}=\min\{M_{\max},P(T+1)\}$. With the default $P=20$ and $M_{\max}=10{,}000$, the cap is reached after approximately $(M_{\max}-P)/P=499$ population iterations, about $15.6$s in this profiling setting. After that, memory remains bounded by replacing old entries.

With the default cap $M_{\max}=10{,}000$, memory lookup took $21.53$s over a mean runtime of $237.97$s, corresponding to $9.05\%$ of total time. As a stress test, increasing the cap to $M_{\max}=100{,}000$ raised lookup time to $139.47$s, or $58.61\%$ of runtime. Thus, exact retrieval is manageable at the memory size used in the main experiments, but becomes a bottleneck for substantially larger archives.

The persistent GPU memory footprint was about $0.09$ GiB for $M_{\max}=10{,}000$ and $0.86$ GiB for $M_{\max}=100{,}000$, including solution vectors and associated one-hot action features. For longer runs or larger memories, approximate retrieval, memory pruning, or reservoir-style memory management are natural extensions.

\subsection{Training the Conditioned Neural Constructive}

We now analyze the behavior of the cNC policy from a bi-objective perspective.
Here we consider cNC in its single-pass greedy form (one construction per call, without the population loop), and study how it trades off solution quality and distance to a conditioning set through the exploration weight~$\omega$ (see Sec.~\ref{subsec:rest}).
We focus on two questions: (i) whether a single conditioned policy can replace a bank of specialized policies trained for fixed trade-offs, and (ii) how the sampling scheme for $\omega$ during training affects the quality of the learned Pareto front.

\begin{figure}[!t]
\centering
\includegraphics[width=0.4\textwidth]{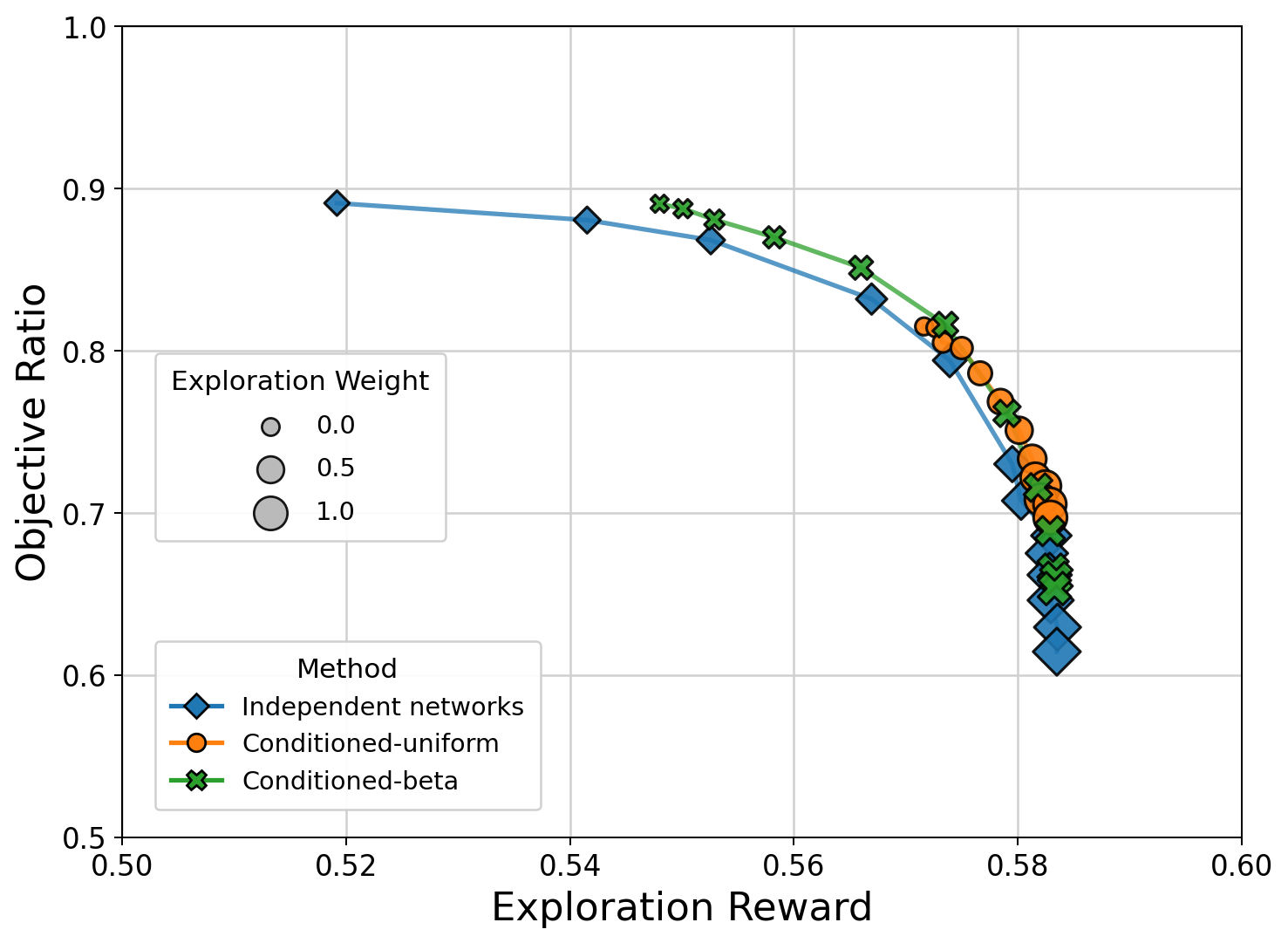}
\caption{Quality--diversity trade-off for independent specialists and conditioned policies. Independent networks are trained separately for fixed scalarization weights, while the conditioned policies use a single network conditioned on the exploration weight $\omega$. We compare conditioned training with uniform sampling $\omega \sim \mathcal{U}[0,1]$ and boundary-biased sampling $\omega \sim \mathrm{Beta}(0.2,0.2)$. Marker size indicates the deployment value of $\omega$, and lines trace the corresponding Pareto-style trajectories.}
\label{fig:quality_diversity_pareto}
\end{figure}

Figure~\ref{fig:quality_diversity_pareto} compares three alternatives: independent networks trained separately for fixed scalarization weights, a conditioned network trained with uniform sampling $\omega \sim \mathcal{U}[0,1]$, and a conditioned network trained with boundary-biased sampling $\omega \sim \mathrm{Beta}(0.2,0.2)$. Each point corresponds to deployment at a different exploration weight $\omega$, with marker size indicating the value of $\omega$.

The conditioned policies recover a competitive quality--diversity frontier with a single parameter vector, avoiding the training and storage cost of maintaining one network per scalarization weight. This also provides continuous inference-time control over the trade-off, rather than restricting deployment to a fixed training grid. Among the conditioned variants, Beta sampling yields a stronger frontier than uniform sampling, especially near the high-quality and high-diversity boundary regions.
This suggests that uniformly sampling $\omega$ can under-emphasize the extreme scalarizations that define the Pareto boundary, whereas concentrating probability mass near $\omega=0$ and $\omega=1$ helps the model learn the endpoints of the trade-off without sacrificing the intermediate regime.

\section{Limitations and Opportunities.} 
\label{sec:discussion}

The proposed methods show that neural constructive and improvement operators can be combined into efficient population-based search procedures through shared memory, conditioning, and restart mechanisms.
At the same time, our implementations are intentionally simple, and they expose a number of practical limitations that are likely to matter as problem size, run length, and population size increase.
This section summarizes the main constraints observed or implied by our design choices, and outlines clear opportunities to improve scalability, better exploit population information, and replace hand-crafted control rules with more adaptive mechanisms.

\textbf{cNI.}
Our cNI implementation relies on an explicit memory that stores many visited solutions.
For very long runs and large populations, keeping all candidates can become prohibitive.
Scalable variants, such as storing only cluster representatives in solution space or applying more aggressive, diversity-aware pruning rules, could preserve coverage of the search space while keeping memory bounded.
Moreover, cNI currently queries memory using a fixed distance function $d(\cdot,\cdot)$ and a weighted average of the $k$ nearest neighbors.
An interesting extension would be to \emph{learn} the retrieval mechanism itself: for example, training an encoder that maps solutions to a latent space where $k$-NN is more informative, and learning attention weights over retrieved neighbors instead of using linear distance decay.

\textbf{cNC.}
The current cNC policy conditions on a fixed-size subset of solutions, encoded as at most $K_{\max}$ additional feature channels. This is sufficient for our purposes, but it inevitably truncates the search history, imposes an arbitrary ordering on the conditioning set (the encoding is not permutation-invariant), and restricts the population structure the model can exploit. A natural line of work is to move toward richer Level~3 conditioning mechanisms: for instance, attending over a compact set of learned memory embeddings rather than raw solutions, using graph encoders that handle variable-size conditioning sets in a permutation-invariant way, or introducing hierarchical memories that summarize different regions of the search space.

\textbf{PB-NCO.}
Our PB-NCO implementation uses a simple patience-based rule to trigger cNC restarts and a hand-crafted schedule for the exploration weight $\omega$.
This mechanism is effective but inherently myopic: it treats all trajectories symmetrically, even though some are more promising to continue than others, and it does not explicitly account for how restarts influence global population diversity.
A natural extension is to replace these rules with adaptive controllers, for example a small learned policy that decides which trajectories to restart based on their recent progress and novelty.

\textbf{Architectural and training heterogeneity.}
All three components rely on a single shared policy trained on a fixed instance distribution.
In practice, population-based methods often benefit from heterogeneity, with different individuals playing distinct roles.
A natural next step is to consider populations of role-conditioned policies tackling the same instance, allowing different policies to specialize on complementary behaviors and thereby exploiting this additional source of diversity.

\textbf{Applicability beyond MC/MIS.}
PB-NCO can in principle be applied beyond MC and MIS, but each new problem requires a suitable solution encoding, distance function, feasible construction or improvement operator, and normalized objective. For binary or set-structured graph problems these ingredients are natural; for routing, assignment, or permutation problems, one would need problem-specific distances, such as edge-set or assignment-based distances, and memory descriptors aligned with the neighborhood structure. Understanding which distances and population representations transfer across problem classes is an important future direction.

\section{Conclusion}
\label{sec:conclusion}

Motivated by the lack of neural counterparts to population-based metaheuristics, this work studied how population structure can be incorporated into neural combinatorial optimization. We evaluated PB-NCO on Maximum Cut and Maximum Independent Set, across ER and RB graph families, large out-of-training graph sizes, five stochastic runs, anytime comparisons, ablations, and sensitivity analyses.

Although population-level coordination introduces additional components and design choices, the empirical gains suggest that learning how candidates interact is a promising direction.
The benefit is strongest when population coordination and learned restarts complement local improvement, especially after improvement-only search begins to plateau.

Overall, the main value of PB-NCO is not only stronger single-solution prediction, but the coordinated allocation of search effort across interacting trajectories. Shared memory provides population-level context for improvement, while cNC restarts introduce new candidates guided by a quality--diversity trade-off. 
We hope this work motivates further research on neural population mechanisms, for example exploring richer communication mechanisms, such as sparse similarity graphs, island models, or attention-based population encoders, together with adaptive diversity-preservation and restart strategies.

\appendices

\section{Problem-Specific Implementation Details}
\label{apendix_problem_implementation}

This appendix summarizes the problem-dependent decoding, normalization, distance, and feature choices used in the experiments.

\subsection{Maximum Cut (MC)}

\textbf{Decoding and symmetry.}
For MC, cNC is implemented in one shot: the network outputs node probabilities $\pi_u\in[0,1]$, and a full cut is obtained by sampling all nodes in a single forward pass. To break the global bit-flip symmetry, we fix one reference node to side $1$ and assign it a dedicated embedding.

\textbf{Reward.}
We use the normalized reward
\[
\tilde f_{\mathcal{I}}(\mathbf{s})
=
\frac{f_{\mathcal{I}}(\mathbf{s})-|\mathcal{E}|/2}
{\sqrt{|\mathcal{E}|\,|\mathcal{V}|\,\log 2/2}},
\]
where $|\mathcal{E}|/2$ is the expected cut value of a random bipartition.

\textbf{Distance and initialization.}
Memory retrieval and cNC's diversity reward use normalized Hamming distance. 
Random initial solutions assign each node independently to one side with probability $1/2$.

\subsection{Maximum Independent Set (MIS)}

\textbf{Feasible decoding.}
For MIS, cNC produces one inclusion score per node in a single forward pass. 
Nodes are then sorted by score and greedily accepted if they are not adjacent to any previously selected node. 
In cNI, feasibility is enforced directly in the action space by masking moves that would violate independence.

\textbf{Decoding-order ablation.}
To isolate the learned cNC ordering from the feasibility projection, we keep the greedy feasibility decoder fixed and vary only the ordering rule.
Table~\ref{tab:mis_decoding_order_ablation} compares random ordering, increasing-degree ordering, dynamic greedy marginal ordering, and learned one-shot cNC orderings trained on ER or RB graphs.

\begin{table}[h]
\centering
\scriptsize
\setlength{\tabcolsep}{4pt}
\renewcommand{\arraystretch}{1.05}
\caption{\textbf{MIS decoding-order ablation.}
All variants use the same greedy feasibility decoder; only the node ordering changes.}
\label{tab:mis_decoding_order_ablation}
\begin{tabular}{@{}lcc@{}}
\toprule
\textbf{Ordering} & \textbf{ER700--800} & \textbf{RB800--1200} \\
\midrule
Random & 29.88 & 30.86 \\
Increasing degree & 33.13 & 34.16 \\
Greedy marginal  (dynamic) & 38.85 & 37.78 \\
Learned cNC (ER-trained) & 38.96 & 32.34 \\
Learned cNC (RB-trained) & 33.52 & 34.37 \\
\bottomrule
\end{tabular}
\end{table}

The results show that the learned ordering contributes beyond the feasibility decoder: the ER-trained cNC ordering slightly improves over greedy marginal ordering on ER graphs and clearly outperforms random and degree-based orderings.
However, transfer across graph families is limited, and on RB graphs the RB-trained cNC ordering improves over random and degree-based orderings but remains below the greedy marginal ordering.
This supports the use of family-matched cNC policies in the main experiments and clarifies that, for MIS, the greedy feasibility decoder alone does not explain cNC performance.

\textbf{Reward.}
Let $L(\mathcal{I})$ be the Caro--Wei lower surrogate \cite{selkow1994probabilistic} and $U(\mathcal{I})$ an upper surrogate based on a greedy maximal matching for instance $\mathcal{I}$.
Given a raw MIS value $f_{\mathcal{I}}(\mathbf{s})$, we normalize the reward as
\begin{equation}
    \tilde f_{\mathcal{I}}(\mathbf{s})
    \;=\;
    \frac{f_{\mathcal{I}}(\mathbf{s}) - L(\mathcal{I})}
         {U(\mathcal{I}) - L(\mathcal{I})}.
\end{equation}
Concretely,
\begin{equation}
    L(\mathcal{I}) \;=\; \sum_{v \in V} \frac{1}{\deg(v)+1},
    \quad
    U(\mathcal{I}) \;=\; N - \mathrm{match}(\mathcal{I}),
\end{equation}
where $\deg(v)$ is the degree of node $v$, $N = |\mathcal{V}|$, and $\mathrm{match}(\mathcal{I})$ is the size of a greedy maximal matching.

\textbf{Distance and initialization.}
As in MC, we use normalized Hamming distance. 
Random feasible solutions are generated by randomized greedy construction.

\subsection{Feature Encoding}
Both problems use edge features derived from the adjacency and node features carrying solution and population information.
For cNI, nodes include the current solution indicator and a memory descriptor obtained by distance-weighted averaging the $k$ nearest stored solutions, following~\cite{garmendia2024marco}.
For cNC, nodes include one binary channel per reference solution in $\mathcal{K}$, up to $K_{\max}$ channels, plus the exploration weight $\omega$ broadcast to all nodes.

\section{Configuration and Neural Network Hyperparameters}
\label{sec:appendix_hyperparameters}

This appendix summarizes the configuration used for the learned components. Table~\ref{tab:training_inference_config} reports the main architectural, training, and inference settings for cNI, cNC, and PB-NCO.

%The architectural hyperparameters in the first row were selected through a controlled study in which we varied the main GT design choices while keeping the training protocol fixed. Each NI model was trained for 10,000 episodes on ER graphs with 20--40 nodes using a batch size of 1024. We then evaluated the final model on 100 ER instances with 60, 100, and 200 nodes for the Maximum Cut problem. Table~\ref{tab:hyperparam_study} reports the average objective values.

\begin{table}[h]
\centering
\scriptsize
\setlength{\tabcolsep}{3pt}
\renewcommand{\arraystretch}{1.08}
\caption{\textbf{Training and inference configuration.}
Main settings used for the learned components. GT denotes Graph Transformer; ER denotes Erd\H{o}s--R\'enyi graphs; RB denotes random binary graphs.}
\label{tab:training_inference_config}
\begin{tabular}{@{}p{0.34\columnwidth}p{0.60\columnwidth}@{}}
\toprule
\textbf{Item} & \textbf{Setting} \\
\midrule
Architecture &
GT encoder; $L{=}3$ for cNI, $L{=}8$ for cNC; $d{=}64$, $h{=}8$, FFNN dim. $256$, GeLU, LayerNorm, dropout $0\%$ \\

cNI training graphs &
ER, $|\mathcal{V}|\in[50,300]$, edge probability $p{=}0.15$ \\

cNC training graphs &
ER, $|\mathcal{V}|\in[50,300]$, $p{=}0.15$; RB, $|\mathcal{V}|\in[200,300]$ \\

Optimizer &
AdamW, lr $=5{\times}10^{-5}$, betas $(0.9,0.95)$, weight decay $0.1$, gradient clipping $1.0$ \\

Training budget &
batch size $32$, population $20$, $100$ epochs $\times$ $100$ episodes \\

Policy-gradient baseline &
PPO with $\gamma{=}0.95$, clip $0.2$, $4$ PPO epochs,
leave-one-out baseline over population \\

Training / evaluation seeds &
$\{1,2,3,4,42\}$ \\

Memory &
$M_{\max}{=}10{,}000$, $k{=}20$ \\

Reward &
cNI repetition penalty $w_{\rm rep}{=}1.0$, cNC exploration weight $\omega\sim\mathrm{Beta}(0.2,0.2)$ \\

Training horizon &
cNI: $2|\mathcal{V}|$ improvement steps per episode; cNC: 20 one-shot rollouts \\

cNC conditioning &
$K_{\max}{=}20$ \\

PB-NCO &
$P{=}20$ individuals, $N_{\rm pat}{=}500$, $\omega_{\rm start}{=}1.0$, $\omega_{\rm end}{=}0.0$, cooling exponent $\phi{=}1.0$\\
\bottomrule
\end{tabular}
\end{table}

\section*{Acknowledgments}
This work was supported by Basque Government through the Elkartek program under the projects KK-2024/00030, KK-2025/00098, KK-2026/00055, KK-2026/00087, and the Research Groups IT2109-26 2026-2029 program; by the Ministry of Science and Innovation MCIN/AEI/10.13039/501100011033 under the projects PID2023-149195NB-I00 and PID2023-152390NB-I00.

\ifCLASSOPTIONcaptionsoff
  \newpage
\fi

\bibliographystyle{IEEEtran}
\bibliography{bibtex/bib/references}

\end{document}